\documentclass[runningheads]{llncs}

\usepackage{eccv}

\usepackage{eccvabbrv}
\usepackage{amsmath,amsfonts,bm}
\usepackage{multirow}
\usepackage{colortbl}
\usepackage{graphicx}
\usepackage{booktabs}
\usepackage{color}
\usepackage[accsupp]{axessibility}  

\usepackage{appendix}
\usepackage{color}
\usepackage[accsupp]{axessibility}  
\usepackage{listings}
\usepackage{xcolor}

\definecolor{codegreen}{rgb}{0,0.6,0}
\definecolor{codegray}{rgb}{0.5,0.5,0.5}
\definecolor{codepurple}{rgb}{0.58,0,0.82}
\definecolor{backcolour}{rgb}{0.95,0.95,0.92}

\lstdefinestyle{mystyle}{
    backgroundcolor=\color{backcolour},   
    commentstyle=\color{codegreen},
    keywordstyle=\color{magenta},
    numberstyle=\tiny\color{codegray},
    stringstyle=\color{codepurple},
    basicstyle=\ttfamily\footnotesize,
    breakatwhitespace=false,         
    breaklines=true,                 
    captionpos=b,                    
    keepspaces=true,                 
    numbers=left,                    
    numbersep=5pt,                  
    showspaces=false,                
    showstringspaces=false,
    showtabs=false,                  
    tabsize=2
}

\usepackage{hyperref}

\usepackage{orcidlink}
\usepackage{marvosym}
\usepackage{ifsym}
\renewcommand\footnotemark{}
\usepackage{wrapfig}
\begin{document}

\title{
Image Compression for Machine and Human Vision with Spatial-Frequency Adaptation
} 

\titlerunning{Adapt-ICMH}

\author{Han Li\inst{1} \and
Shaohui Li\inst{2}\textsuperscript{(\Letter)} \and
Shuangrui Ding\inst{3} \and
Wenrui Dai \inst{1}\textsuperscript{(\Letter)} \and \\
Maida Cao\inst{1}
 \and
Chenglin Li\inst{1}
 \and
Junni Zou\inst{1}
 \and
Hongkai Xiong\inst{1}\thanks{Correspondence to Wenrui Dai and Shaohui Li.}}

\authorrunning{H. Li et al.}

\institute{Shanghai Jiao Tong University \and
Tsinghua Shenzhen International Graduate School, Tsinghua University \and 
The Chinese University of Hong Kong
\\
\email{\{qingshi9974,daiwenrui,caomaida,lcl1985,zoujunni,xionghongkai\}@sjtu.edu.cn} 
\email{lishaohui@sz.tsinghua.edu.cn},~~\email{ds023@ie.cuhk.edu.hk}
}

\maketitle

\begin{abstract}
Image compression for machine and human vision (ICMH) has gained increasing attention in recent years. Existing ICMH methods are limited by high training and storage overheads due to heavy design of task-specific networks. To address this issue, in this paper, we develop a novel lightweight adapter-based tuning framework for ICMH, named Adapt-ICMH, that better balances task performance and bitrates with reduced overheads. We propose a spatial-frequency modulation adapter (SFMA) that simultaneously eliminates non-semantic redundancy with a spatial modulation adapter, and enhances task-relevant frequency components and suppresses task-irrelevant frequency components with a frequency modulation adapter. 
The proposed adapter is plug-and-play and compatible with almost all existing learned image compression models without compromising the performance of pre-trained models. Experiments demonstrate that Adapt-ICMH consistently outperforms existing ICMH frameworks on various machine vision tasks with fewer fine-tuned parameters and reduced computational complexity. Code will be released at \url{https://github.com/qingshi9974/ECCV2024-AdpatICMH}.

\end{abstract}

\section{Introduction}
\label{sec:intro}

\begin{figure}[tb]
  \centering
  \includegraphics[width=1\linewidth]{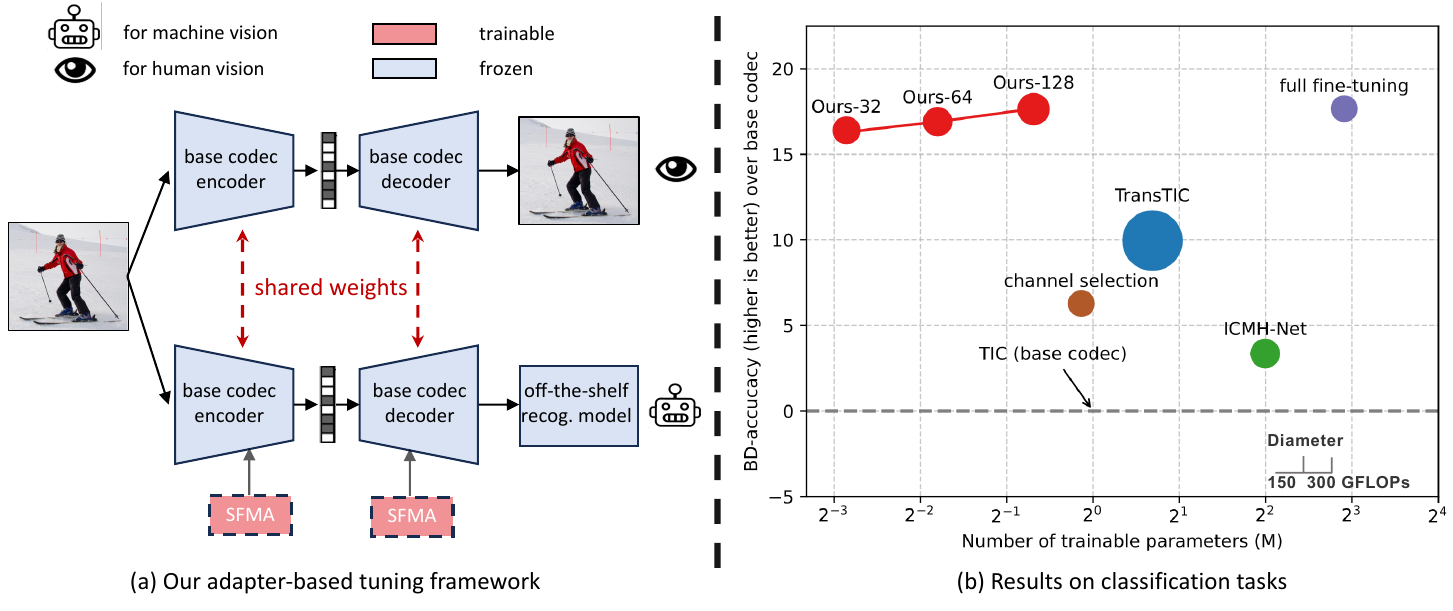}
  \caption{\textbf{Left:}  our adapter-based tuning framework. \textbf{Right:} Rate-accuracy performance comparison on classification for ImageNet-\textit{val}~\cite{deng2009imagenet}. We compare our methods (Ours-$n$ indicates $n$ middle dimensions for SFMA) with full fine-tuning, TransTIC~\cite{chen2023transtic}, ICMH-Net~\cite{liu2023icmh}, and channel selection~\cite{liu2022improving}. BD-accuracy is computed by replacing the PSNR in BD-PSNR~\cite{bdrate} with top-1 accuracy and setting the base codec of TIC~\cite{lu2022transformer} as the anchor.  The size of circles indicates GFLOPs for inference during encoding.}
  \label{fig:overall}
\end{figure}

The booming of computer vision~\cite{he2016deep,he2017mask,ren2015faster,liu2021swin,sun2019deep,xie2021segformer,ding2022motion,zheng2024bem} raises the demands for excessive images serving machine vision systems and accomplishing tasks such as classification, detection, segmentation. In practice, images from users have to be compressed and transmitted to exploit the off-the-shelf recognition models deployed in the cloud. However, compression directly using existing image codecs, especially learned image compression (LIC) models~\cite{balle2016end,balle2018variational,minnen2018joint,minnen2020channel,zou2022devil,liu2023learned,li2023frequency} optimized for human vision, could cause the loss of semantic information contained in images and dramatically reduce the accuracy of downstream tasks. To meet the demands of both machine and human vision systems, unified codecs to achieve image compression for machine and human vision (ICMH) have been explored~\cite{choi2022scalable,codevilla2021learned,liu2022improving,bai2022towards,chen2023transtic,liu2023icmh,fischer2022boosting,feng2023semantically,liu2021semantics,yang2021towards,liu2024rate,liu2023composable}.

A multi-task pipeline has been built for ICMH in~\cite{choi2022scalable,codevilla2021learned,bai2022towards}, where a task-agnostic encoder and multiple task-specific decoders are \textbf{jointly trained from scratch} for human vision and multiple machine vision tasks. However, the multi-task pipeline results in degraded rate-distortion (R-D) performance for human vision and is restricted in adapting to newly coming machine vision tasks.


Since LIC models are usually fitted to the distributions of natural images~\cite{lee2018context,mentzer2019practical}, recent works~\cite{chen2023transtic,liu2023icmh,liu2022improving,fischer2022boosting,liu2023composable} tend to adopt a tuning framework that adapts a \textbf{pre-trained} image codec optimized for human vision (called base codec) to diverse machine vision tasks without sacrificing R-D performance. This pipeline enables ICMH by training only a lightweight module specifically for the machine vision task, while freezing the base codec parameters for sharing across human and machine vision.
In ICMH-Net~\cite{liu2023icmh}, a spatial-channel mask generator is trained to determine a subset of quantized latent for a specific machine vision task. Liu~\emph{et al.}~\cite{chen2023transtic} utilized visual prompt tuning~\cite{jia2022visual}  to transfer the transformer-based base codec~\cite{lu2022transformer} to machine vision tasks. Unfortunately, these methods suffer from evidently degraded performance compared to codecs fully fine-tuned for machine vision tasks. Besides, they introduce significant overheads of computational and model complexity to the base codec due to their task-specific designs. \emph{It remains a challenge to achieve efficient fine-tuning of pre-trained human vision-oriented image codecs on machine vision tasks.}

In this paper, we consider the fact hat almost all spatial and frequency components of an image are exploited by the base codec to improve the visual quality for human vision but could be redundant for machine vision tasks. We are thereby motivated to modulate the intermediate feature of the base codec in both spatial and frequency domains such that the redundancy in the latent representation can be reduced and the reconstructed image can be efficiently adapted to various machine vision tasks.

To this end, we propose a play-and-plug module of spatial-frequency modulation adapter (SFMA) to better balance the machine vision task performance and transmission cost. 
Specifically, the proposed SFMA consists of two parallel submodules, \emph{i.e.}, a spatial modulation adapter (SMA) that eliminates non-semantic redundancies in the spatial domain and a frequency modulation adapter (FMA) that amplifies the frequency components that contribute more to the machine vision task ($e.g.,$ high-frequency components are more crucial for segmentation) and suppresses the redundant frequency components. Furthermore, we develop an adapter-based tuning framework for ICHM, named Adapt-ICMH, by plugging the proposed SFMA into both the encoder and decoder of the base codec, as depicted in \cref{fig:overall}. It should be noted that, different from~\cite{chen2023transtic} that can only adapt the \textit{transformer-based} base codec, Adapt-ICMH is \textit{architecture-agnostic} such that it is compatible with almost all existing LIC models for adaptation to various machine vision tasks.
In summary, our contributions include:
\begin{itemize}
\item We propose a spatial-frequency modulation adapter (SFMA) to efficiently update the intermediate feature and better balance the performance and bitrates for diverse machine vision tasks.

\item Based on SFMA, we develop a novel tuning framework named Adapt-ICMH to achieve image coding for machine and human vision with a shared base codec,  significantly reducing the training and storage overhead.

\item Our SFMA is plug-and-play, which could be incorporated with all existing LIC models to achieve ICMH. 
Experiments show that our method consistently outperforms other ICMH methods in various machine vision tasks with even reduced model and computational complexity.

 \end{itemize}
\section{Related Work}
\textbf{Learned Image Compression} (LIC) has attracted wide attention due to its superior rate-distortion (R-D) performance. Research in LIC can be categorized into two main paths. The first focuses on enhancing the nonlinear transforms~\cite{balle2020nonlinear}, evolving from convolutional neural networks (CNN)-based~\cite{balle2016end,balle2018variational,minnen2018joint,minnen2020channel}
to attention-based transforms~\cite{cheng2020learned}, and more recently, transformer-based transforms~\cite{lu2022transformer,zou2022devil,liu2023learned,li2023frequency}. The second path involves designing more powerful entropy models. Prior works~\cite{balle2016end,balle2018variational} proposed factorized and hyperprior models, while recent efforts~\cite{minnen2018joint,minnen2020channel,koyuncu2022contextformer,qian2021entroformer} tend to introduce spatial or channel-wise autoregression in the entropy model. However, these methods focus solely on the human visual quality of the decoded image, while ignoring the application in downstream machine vision tasks.

\noindent\textbf{Image Compression for Machine and Human Vision} (ICMH)  has become a hot topic in recent years due to the growing number of images that need to be processed for various machine vision tasks, such as classification~\cite{deng2009imagenet,he2016deep,huang2017densely} , detection~\cite{carion2020end,zhu2020deformable,girshick2015fast}, segmentation~\cite{he2017mask,strudel2021segmenter,zheng2021rethinking,shi2022transformer} and pose estimation~\cite{sun2019deep,cao2017realtime,li2021hierarchical,li2023pose}.  One straight way is to fully fine-tune the base codec on machine vision tasks. However, this requires storing and deploying a separate copy of the entire codec network for each individual task, which severely hampers practical applicability.
 
Earlier works~\cite{choi2022scalable,codevilla2021learned,liu2022improving,bai2022towards} presented a multi-task pipeline to unify the image coding for machine and human vision. However, it still requires multiple task-specific decoders, leading to increased storage overhead. In addition, the entire network must be trained from scratch, resulting 
in significant training overhead and impeding the adaptation to newly coming machine vision tasks.  To address this problem, Chen~\emph{et al.}~\cite{chen2023transtic} proposed to leverage the pre-trained human vision-oriented base codec and train an extra spatial-channel mask generator to transmit a subset of latent for machine vision task. Unfortunately, it cannot be easily extended to autoregression-based entropy models~\cite{minnen2018joint,minnen2020channel} and fails to achieve adaptation of the nonlinear transforms.  Liu~\emph{et al.}~\cite{liu2023icmh} proposed a prompt-based tuning framework that introduces prompts in the transformer-based nonlinear transforms for machine vision. However, it cannot be compatible with the mainstream CNN-based LIC methods~\cite{balle2016end,balle2018variational,cheng2020learned} and the performance is still highly inferior to full fine-tuning.  In contrast, our framework generalizes to almost all existing LIC models with different nonlinear transforms and entropy models and achieves comparable rate-accuracy performance compared to full fine-tuning. See Appendix B for more discussions about ICMH.

\noindent\textbf{Parameter-Efficient Fine-Tuning} (PEFT) has been extensively studied for NLP~\cite{houlsby2019parameter,pfeiffer2020adapterhub,pfeiffer2020adapterfusion,he2021towards,lester2021power,liu2023pre}.  It can effectively fine-tune a pre-trained large-scale model on downstream tasks or datasets with the major model frozen.
With the advent of vision transformers~\cite{dosovitskiy2020image,liu2021swin,carion2020end}, researchers have applied PEFT tools to machine vision tasks, including prompt tuning~\cite{jia2022visual,khattak2023maple}, adapters~\cite{chen2022adaptformer,chen2022vision,wang2023adapting}, and LoRA~\cite{he2023parameter}. In the field of learned image compression,
Feng~\emph{et al.}~\cite{feng2023prompt} first introduced prompts for machine-oriented image coding, while Liu~\emph{et al.}~\cite{chen2023transtic} further explored the application of prompt-tuning for ICMH. ~\cite{lv2023dynamic,shen2023dec,tsubota2023universal}  transmitted the parameters of additional adapters for content-adaptive optimization~\cite{lee2018context,campos2019content} to improve the human visual quality of the decoded image, which is completely different from ours in terms of methodology and application scenarios. In addition, naive adapters~\cite{liu2023composable,houlsby2019parameter,chen2022adaptformer} update the feature without considering the transmission cost of different underlying frequencies and spatial components of the image, which is vital for ICMH. In our paper, we propose a novel spatial-frequency modulation adapter specifically for ICHM to address this limitation.
\section{Methods}

\subsection{Empirical Findings by Full Fine-tuning}

\begin{figure}[tb]
  \centering
  \includegraphics[width=0.96\linewidth]{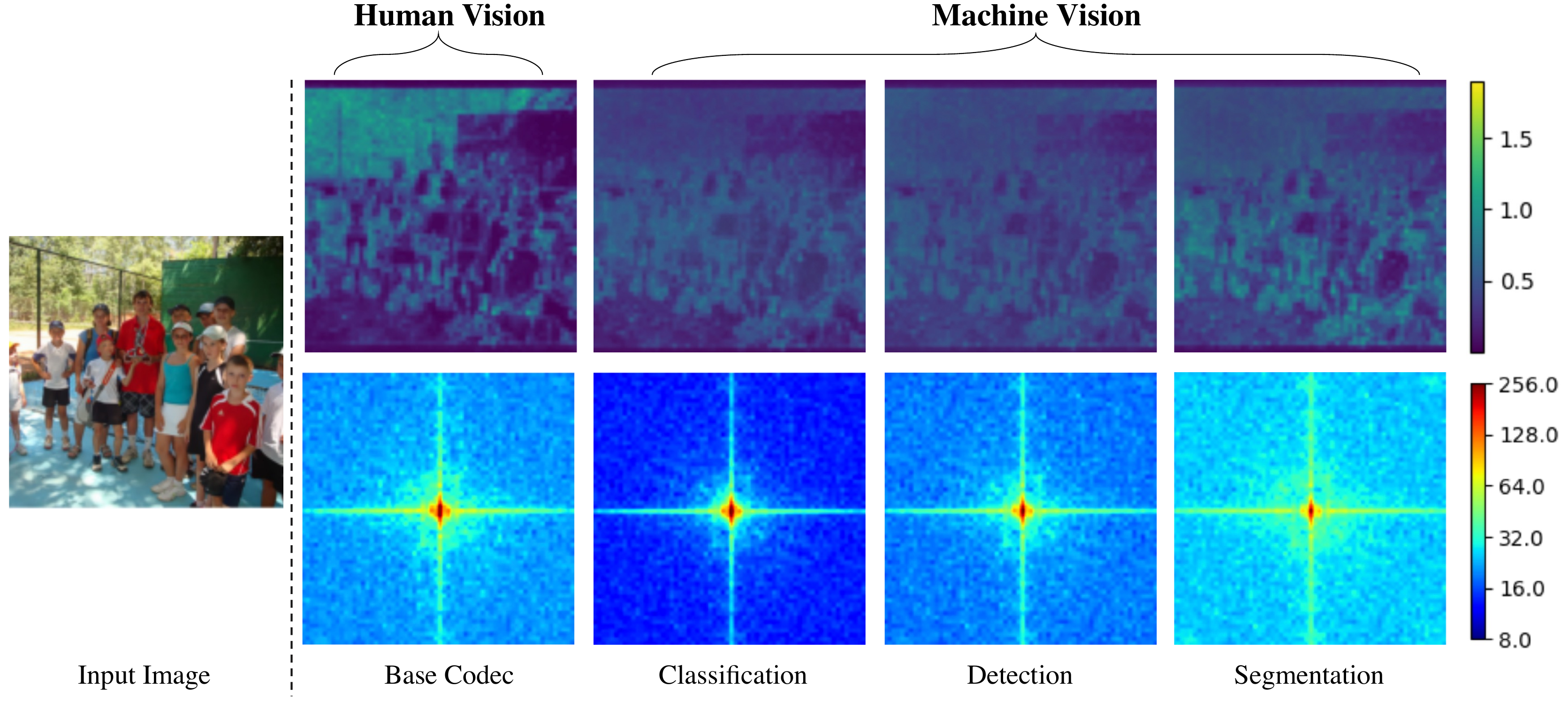}
  \caption{Visualization of the bit allocation maps (first row) and power spectral density maps (second row) of the latent $\hat{y}$. The left part shows the raw input image. Each column of the right part corresponds to a codec for each task, including the base codec for human vision and three fine-tuned codecs for machine vision tasks. The bit allocation map is computed by averaging the negative log-likelihood (\textit{i.e.,} $-\log_2p(\hat{y})$) across channels. The power spectral density map is computed by applying the Fast Fourier Transform (FFT) to $\hat{y}$ with a shift operation to center the zero frequency component, and then averaging its absolute value across channels.}
  \label{fig:motivation}
\end{figure}

We first conduct a simple experiment to highlight the key insight of this paper:
 image compression for machine vision tends to prioritize transmitting \textit{distinct spatial and frequency information} compared to human vision. This preference also varies across specific machine vision tasks.

Taking the pre-trained mean-scale hyperprior\footnote{We use the pre-trained model (named \textit{mbt2018-mean}) offered by CompressAI~\cite{begaint2020compressai}.} model~\cite{minnen2018joint} optimized for human vision as the base codec, we fully fine-tune\footnote{We use loss function \eqref{eq:loss-function} to conduct the full fine-tuning.} it on three different machine vision tasks, including classification, object detection, and instance segmentation.  Then, we feed the same image to the base codec and three fine-tuned codecs, respectively.  The bit allocation map and power spectral density map of latent $\hat{y}$ for each codec are shown in \cref{fig:motivation}.

From the bit allocation map, we can see that the base codec consumes more bitrates in the complex background region, while the codecs that are fully fine-tuned on machine  tasks pay more attention to the semantic object.
In addition, we can see from the power spectral density maps that the base codec tends to encode frequency components as many as possible to achieve higher visual quality. However, many of these frequency components are redundant for machine vision. For example, instance segmentation prefers more mid- and high-frequency components for accurate edge and contour reconstruction, while low- and mid-frequency components are sufficient for simpler tasks  (\textit{e.g.,} classification).

These observations confirm that image compression has different spatial and frequency domain preferences for different tasks. However, fully fine-tuning the base codec on different machine vision tasks is costly in terms of training time and memory. Thus, in this paper, we propose a spatial-frequency modulation adapter to conduct efficient fine-tuning, and effectively reduce the redundancies in both spatial and frequency domains for machine vision tasks.

\subsection{Framework Overview}
\begin{figure}[htb]
  \centering
  \includegraphics[width=1\linewidth]{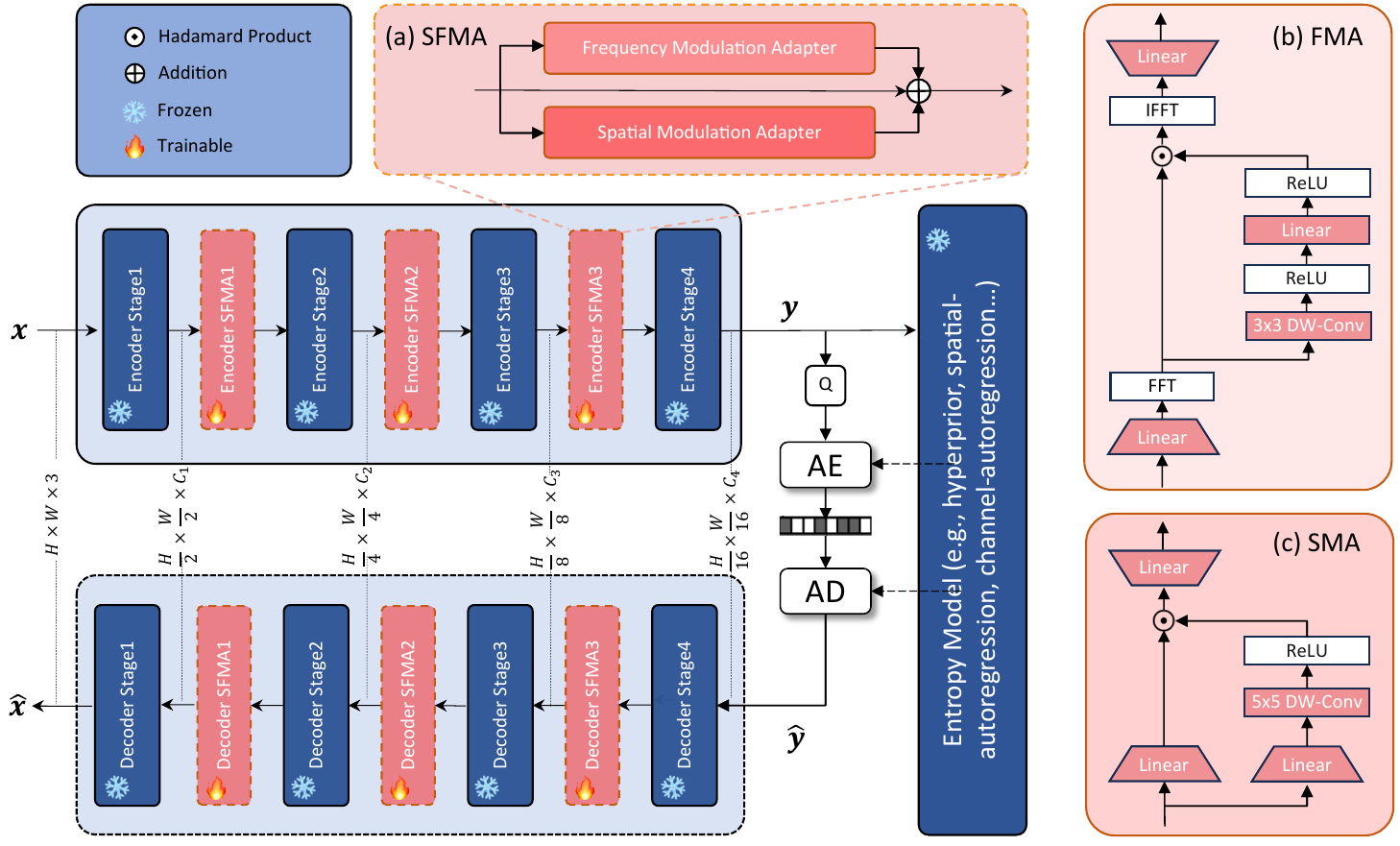}
  \caption{Overview of our proposed Adapt-ICMH framework. Multiple spatial-frequency modulation adapters (SFMA) are plugged into the encoder $g_a$ and decoder $g_s$ of the base codec for feature adaptation. During the adaptation to the machine vision task, the base codec is frozen and only these adapters are trainable. 
  For briefness, we do not illustrate the specific architecture of the encoder, decoder stage, and entropy model, as it depends on the specific base codec. Please see Appendix E for the detailed architecture.}
  \label{fig:overview}
\end{figure}
Our compression framework Adapt-ICMH, shown in \cref{fig:overview}, is designed to adapt the existing learned image compression model that is optimized for human vision (\textit{i.e.,} base codec) to downstream machine vision tasks.  To this end, we propose the Spatial-Frequency Modulation Adapter (SFMA), a plug-and-play bottleneck module, which is plugged into the encoder $g_a$ and decoder $g_s$ of the base codec for feature adaptation. 

During the adaptation stage, the weights of the base codec (\textcolor{blue}{blue part}) are loaded from the pre-trained checkpoint and kept frozen.  Therefore, we only optimize the added adapters (\textcolor{red}{red part}) for specific machine vision tasks with the following loss function:

\begin{equation} 
    \mathcal{L}= \mathcal{R}+\lambda \cdot \mathcal{D}(\bm{x}, \hat{\bm{x}}; \mathcal{G}),
    \label{eq:loss-function}
\end{equation}
where $\mathcal{R}$ denotes the overall estimated bitrates, $\mathcal{D}$ calculates the task-specific perceptual distortion between raw image $\bm{x}$ and reconstructed image $\hat{\bm{x}}$ with an off-the-shelf recognition model $\mathcal{G}$ (\textit{e.g.}, Faster RCNN~\cite{ren2015faster} for object detection). $\lambda$ is a trade-off term to balance the task performance and bitrates. Please refer to Appendix C for more details about loss function. 

After adaptation, we can only modify the weights of lightweight adapters for different machine vision tasks, while keeping the parameters of the base codec unchanged. If an image needs to be compressed for human vision, we can directly remove the adapters to obtain the original base codec, so that the R-D performance for human vision is not affected. In this way, a single base codec model can be used for both human vision and multiple machine vision tasks.

\subsection{Spatial-Frequency Modulation Adapter}
The proposed Spatial-Frequency Modulation Adapter (SFMA) consists of a Spatial  Modulation Adapter (SMA) and a Frequency Modulation Adapter (FMA), which are configured in a parallel way. As shown in \cref{fig:overview}, 
 the encoder $g_a$ of base codec generally consists of four stages, denoted by $g_a = g_{a1}\circ g_{a2}\circ g_{a3}\circ g_{a4}$.  Given the input image $\bm{x}\in\mathbb{R}^{3\times H\times W}$, where $H\times W$ is the spatial resolution, we define the output feature of the  $j$-th intermediate stage of $g_a$ as $\bm{x}_j\in\mathbb{R}^{C_j\times H_j \times W_j}$, where $C_j$ is the number of channels, $H_j =H/2^j$, $W_j=W/2^j$, and $j \in \left\{1,2,3\right\}$. Supposing that the SFMA plugged after the $j$-th encoder stage is denoted by $\operatorname{SFMA}_j$, we can obtain the adapted feature $\tilde{\bm{x}}_j$ by aggregating original input feature $\bm{x}_j$ with the  adapted component. The process can be defined as follows:
\begin{equation}
\begin{aligned}
\tilde{\bm{x}}_j &= \operatorname{SFMA}_j (\bm{x}_j) \\
&= \bm{x}_j + \underbrace{\operatorname{FMA}_j(\bm{x}_j)+ \operatorname{SMA}_j(\bm{x}_j)}_{\text{adapted component}},
\end{aligned}
\end{equation}
where $\operatorname{FMA}_j$ and $\operatorname{SMA}_j$ denote the $j$-th frequency and spatial modulation adapter, respectively. The process of the adapter for decoder $g_s$ can be derived similarly.

\noindent\textbf{Frequency Modulation Adapter.} Our frequency modulation adapter (FMA) is designed to adapt the feature in the frequency domain. By this modulation operation,  we can eliminate frequency redundancies and amplify the important frequency components for machine vision tasks. 
Specifically, the input $\bm{x}_j \in\mathbb{R}^{C_j\times H_j \times W_j}$ firstly be down-projected to a bottleneck middle dimension $\hat{C}$ ($\hat{C}\le C_j$) by a linear layer $\bm{W}_{\text{down}}^\text{f}\in\mathbb{R}^{C_j\times \hat{C}}$ and then be transformed into frequency domain by fast Fourier transform (FFT), obtaining the middle feature $\bm{x}^\text{f}_j$. Subsequently, we generate a frequency modulation matrix $\bm{m}^\text{f}_j$ and then use Hadamard product to calculate the modulated result $\bar{\bm{x}}^\text{f}_j$. Finally, $\bar{\bm{x}}^\text{f}_j$ is inversely transformed using inverse FFT (IFFT) and up-projected by another linear layer $\bm{W}_{\text{up}}^\text{f}\in\mathbb{R}^{\hat{C}\times C_j}$.  The overall process of FMA can be formulated as follows:
\begin{equation}
\begin{aligned}
\bm{x}^\text{f}_j = & \mathcal{F} (\bm{x}_j\cdot\bm{W}_{\text{down}}^\text{f}),\\
\bm{m}^\text{f}_j = &\sigma(\operatorname{DW-Conv}_{3\times3}(\bm{x}^\text{f}_j))\cdot\bm{W}_{\text{middle}}^\text{f},\\
\bar{\bm{x}}^\text{f}_j = &\bm{x}^\text{f}_j \odot \sigma(\bm{m}^\text{f}_j ),\\
\tilde{\bm{x}}^\text{f}_j = & \mathcal{F}^{-1}(\bar{\bm{x}}^\text{f}_j) \cdot\bm{W}_{\text{up}}^\text{f}
\end{aligned}
\end{equation}
where $\sigma$ denotes ReLU~\cite{nair2010rectified}, $\operatorname{DW-Conv}_{3\times3}$ denotes depth-wise convolution~\cite{chollet2017xception} with kernel size $3\times 3$, $\bm{W}_{\text{middle}}^\text{f}\in\mathbb{R}^{ \hat{C}\times \hat{C}}$ is a  linear layer, $\odot$ denotes Hadamard
product, and $\mathcal{F}(\cdot)$ and $\mathcal{F}^{-1}(\cdot)$ denote the FFT and IFFT, respectively.

\noindent\textbf{Spatial Modulation Adapter.} Our spatial modulation adapter (SMA) is designed to adapt features in the spatial domain, thus guiding the codec to focus more on the semantic region and reduce the spatial redundancies in the background.  Similarly, the input $\bm{x}_j \in\mathbb{R}^{C_j\times H_j \times W_j}$ is first down-projected to a bottleneck middle dimension $\hat{C}$ by a linear layer $\bm{W}_{\text{down1}}^\text{s}\in\mathbb{R}^{C_j\times \hat{C}}$, obtaining the middle feature $\bm{x}^\text{s}_j$. Then, we produce the spatial modulation matrix $\bm{m}^\text{s}_j$ by feeding $\bm{x}_j $ to another down-projection linear layer $\bm{W}_{\text{down2}}^\text{s}\in\mathbb{R}^{C_j\times \hat{C}}$ followed by a $5\times 5$ depth-wise convolution layer. We then apply an up-projection linear layer $\bm{W}_{\text{up}}^\text{s}\in\mathbb{R}^{\hat{C}\times C_j}$ to the modulated feature to get the spatial adapted result.  The overall process of SMA can be formulated as follows:

\begin{equation}
\begin{aligned}
\bm{x}^\text{s}_j = & \bm{x}_j\cdot\bm{W}_{\text{down1}}^\text{s},\\
\bm{m}^\text{s}_j = &\operatorname{DW-Conv}_{5\times5}(\bm{x}^\text{s}_j\cdot\bm{W}_{\text{down2}}^\text{s}),\\
\bar{\bm{x}}^\text{s}_j = &\bm{x}^\text{s}_j \odot \sigma(\bm{m}^\text{s}_j ),\\
\tilde{\bm{x}}^\text{s}_j = & \bar{\bm{x}}^\text{s}_j\cdot\bm{W}_{\text{up}}^\text{s},
\end{aligned}
\end{equation}

\section{Experiments}

\subsection{Training Details and Datasets.}
To demonstrate the generalization ability of our framework, we adopt three mainstream LIC methods, \emph{i.e.,} stacked convolution-based~\cite{minnen2018joint}, residual network-based ~\cite{cheng2020learned}, and transformer-based~\cite{lu2022transformer} image compression model. We use their pre-trained models from \textit{CompressAI}~\cite{begaint2020compressai} library as base codes, denoted by \emph{mbt2018-mean}, \emph{cheng2020-anchor} and \emph{Lu2022-TIC}, respectively. 
For benchmark machine vision tasks, we evaluate our framework on classification, object detection, and instance segmentation, respectively. 
By freezing these base codecs pre-trained for human vision, we then train our SFMAs using the task-specific loss of Eq.~\eqref{eq:loss-function} with the bottleneck middle dimension $\hat{C}$ set to 64.  
Specifically, we train the classification task on ImageNet-\emph{train} dataset~\cite{deng2009imagenet} for 8 epochs with a batch size of 16, and train the object detection and instance segmentation tasks on COCO2017-\emph{train} dataset~\cite{lin2014microsoft} both for 40 epochs with a batch size of 8. The images are randomly cropped and resized to $256\times256$ for training all tasks. 
For the calculation of task-specific perceptual distortion term in the Eq.~\eqref{eq:loss-function}, we follow~\cite{chen2023transtic} to use the pre-trained ResNet50~\cite{he2016deep}, Faster R-CNN~\cite{ren2015faster} and Mask R-CNN~\cite{he2017mask} for classification, object detection, and instance segmentation, respectively. More training details are included in Appendix C.

\subsection{Evaluation}
We follow the setting of~\cite{chen2023transtic} to evaluate the performance of our method. Specifically,
we use bits per pixel (BPP) to evaluate bitrates for compression.  For image classification,
the evaluations are conducted on the ImageNet-\emph{val} dataset~\cite{deng2009imagenet} and images are resized and center cropped to $256\times256$. We adopt the pre-trained ResNet50 from \emph{torchvision} library as the off-the-shelf recognition model and use  top-1 accuracy as the quality metric.
For objection detection and instance segmentation, the evaluations are both done on the COCO2017-\emph{val} dataset~\cite{lin2014microsoft} by using the pre-trained Faster R-CNN and Mask R-CNN from \emph{detectron2} library as the off-the-shelf recognition model, respectively. The test images are resized to 800 pixels for the short side and we use mean average precision (mAP) with an Intersection of Union (IoU) threshold of 0.5 as the quality metric.

\subsection{Rate-Accuracy Comparison}
We compare the rate-accuracy performance of our methods with the state-of-the-art (SOTA) methods\footnote{
Since ICMH-Net~\cite{liu2023icmh} and channel selection~\cite{liu2022improving} are not open source and use different recognition models and test datasets than ours, we reproduce them with the same training settings as ours.  See Appendix D for more details.}, including \emph{TransTIC}~\cite{chen2023transtic}, \emph{ICMH-Net}~\cite{liu2023icmh}, and \emph{channel selection}~\cite{liu2022improving}. Note that our method has an equal R-D performance for human vision with these methods, since we all adopt the pre-trained image codec as base codec and freeze it.  We also evaluate the rate-accuracy performance of the \emph{base codec} and further \textit{full fine-tuning} it on machine vision tasks end-to-end.

\begin{figure*}[t]
\begin{center}
\subfloat[\textit{Lu2022-TIC} model~\cite{lu2022transformer} as base codec]{\includegraphics[width=.95\linewidth]{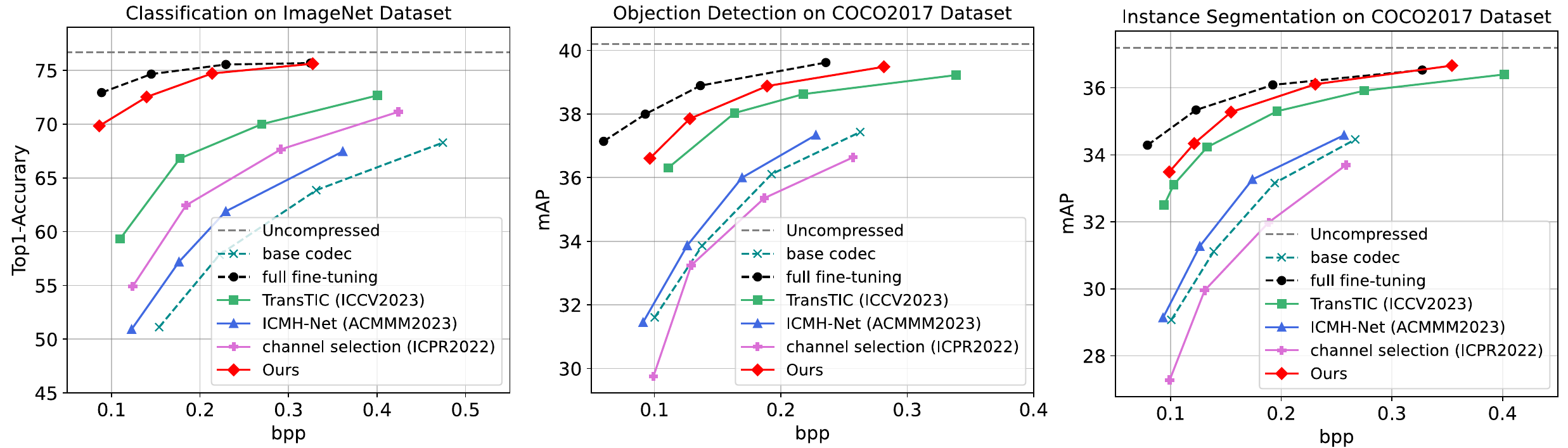}}\\
\subfloat[\textit{Cheng2020-anchor} model~\cite{cheng2020learned} as base codec]{\includegraphics[width=.95\linewidth]{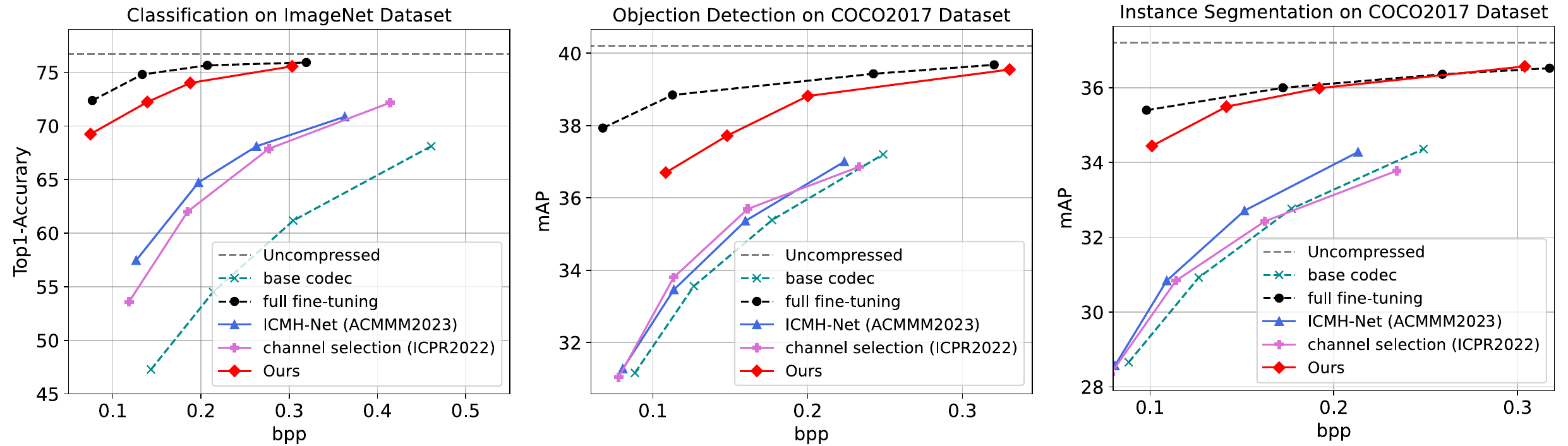}}\\
\subfloat[\textit{mbt2018-mean} model~\cite{minnen2018joint} as base codec]{\includegraphics[width=.95\linewidth]{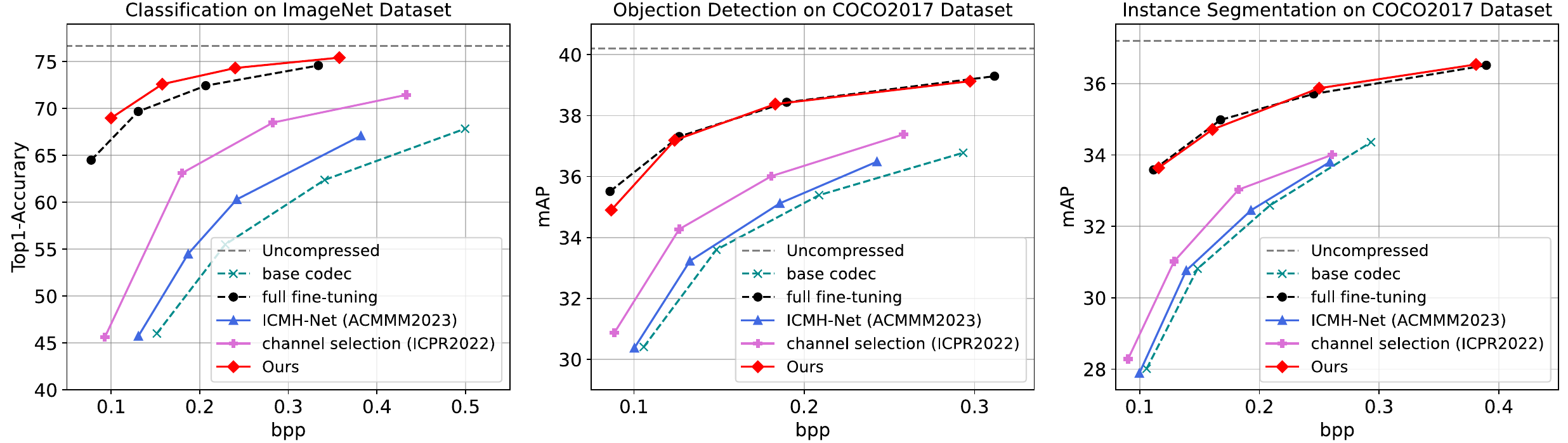}}
\caption{Rate-Accuracy performance comparison under different machine vision tasks and different base codecs.}
\label{fig:main_results}
\end{center}
\end{figure*}

The rate-accuracy curves for competing methods are shown \cref{fig:main_results}. In addition,
we also summarize the average bitrates savings, accuracy improvements, and the trainable parameters count of different methods in \cref{tab:main_results}. Specifically, we report the BD-rate~\cite{bdrate} results to quantify the average bitrates savings with equal task accuracy, we also follow~\cite{chen2023transtic} to report BD-acc/BD-mAP results to measure the average task accuracy improvements with equal bitrates.  BD-acc/BD-mAP is calculated by replacing the PSNR in BD-PSNR  with top-1 accuracy/mAP.

From the results of \cref{fig:main_results} and \cref{tab:main_results}, we can observe that \textbf{(1)} our method consistently achieves excellent performance across all three machine vision tasks and outperforms~\cite{chen2023transtic,liu2023icmh,liu2022improving}, the superiority is more significant on the classification task; \textbf{(2)}  Our method demonstrates strong generalization ability as it can effectively generalize to three different base codecs, despite their varying architectures of nonlinear transforms (i.e., $g_a$ and $g_s$) and entropy models. \textbf{(3)}  our method achieve comparable or even superior performance to \textit{full fine-tuning}. However, \textit{full fine-tuning} requires storing and deploying a separate copy of the entire codec for each task, significantly increasing training and storage overhead and severely hindering practical applicability. \textbf{(4)}  our method requires fewer trainable parameters (\emph{i.e.}, about $1\sim5\%$ of the number of parameters of the base codec) compared to other tuning-based methods (\emph{i.e.},  about $1\sim5\%$ of the number of parameters of the base codec), which demonstrates the parameter efficiency of our method.  For the application on larger transformer-based image codecs~\cite{zou2022devil,liu2023learned,li2023frequency}, see Appendix H.
For the comparison of computational complexity, see Appendix G.

\begin{table}[htp]
\centering
\caption{Comparison of rate-accuracy performance and the number of trainable parameters under different machine vision tasks and different base codecs. BD-rate, BD-acc, BD-mAP are used to quantitatively measure the rate-accuracy performance with the base codec as anchor, respectively. Arrows indicate whether lower is better ($\downarrow$) or higher is better ($\uparrow$). (-) indicates the BD-rate can not be computed.}
\resizebox{1\textwidth}{!}{\begin{tabular}{l|rr|rr|rr|r}
\hline
  \multirow{2}{*}{\textbf{Method}} & \multicolumn{2}{c|}{\textbf{Classification}} & \multicolumn{2}{c|}{\textbf{Detection}} & \multicolumn{2}{c|}{\textbf{Segmentation}}& Trainable \\
         & BD-rate$\downarrow$ & ~BD-acc$\uparrow$~ & BD-rate$\downarrow$ & BD-mAP$\uparrow$ & BD-rate$\downarrow$ & BD-mAP$\uparrow$ &Params $\downarrow$(M) \\
\hline
\multicolumn{8}{c}{\textit{Lu2022-TIC}} \\
\hline
 \rowcolor{gray!20} 
 full fine-tuning & -&17.68&-73.94\%&4.51&-67.98\%&3.76&7.51(100\%)\\
 TransTIC~\cite{chen2023transtic}  &-58.32\%&9.96&-46.30\%&2.77&-46.41\%&2.70&1.61(21.4\%)\\
 ICMH-Net~\cite{liu2023icmh}   &-18.75\%&3.36&-9.07\%&0.63&-10.77\%&0.65&3.98(53.0\%)\\
 channel selection~\cite{liu2022improving} & -37.17\%&6.28&6.84\%&-0.55&16.51\%&-0.94&0.91(12.1\%)\\
 Ours  &-88.57\%&16.90&-55.14\%&3.55&-52.40\%&3.21&0.28(3.7\%)\\
\hline
\multicolumn{8}{c}{\textit{Cheng2020-anchor}} \\
\hline
 \rowcolor{gray!20} 
 full fine-tuning  &-&21.04&-59.20\%&4.69&-76.03\%&3.86&26.60(100\%)\\
 ICMH-Net~\cite{liu2023icmh}  &-47.46\%&10.4&-8.81\%&0.53&-12.18\%&0.74&4.43(16.6\%)\\
  channel selection~\cite{liu2022improving} & -41.58\%&8.77&-11.66\%&0.73&5.22\%&0.22&1.34(4.8\%)\\
 Ours  &-87.56\%&20.27&-49.34\%&3.13&-59.90\%&3.48 &0.41(1.5\%)\\
\hline
\multicolumn{8}{c}{\textit{mbt2018-mean}}\\
\hline
 \rowcolor{gray!20} 
 full fine-tuning & -79.92\%&17.97&-60.87\%&3.87&-56.64\%&3.28&7.03(100\%)\\
 ICMH-Net~\cite{liu2023icmh}  & -15.99\%&3.55&-6.02\%&0.39&-4.97\%&0.31&3.98(56.6\%)\\
channel selection~\cite{liu2022improving} & -41.30\%&9.90&-23.07\%&1.52&-15.09\%&1.05&0.91(12.9\%)\\
 Ours  &-82.00\%&18.71&-56.17\%&3.84&-52.65\%&3.17&0.28(4.1\%)\\
\hline
\end{tabular}}
\label{tab:main_results}
\end{table}
\begin{figure}[h]
  \centering
  \includegraphics[width=1\linewidth]{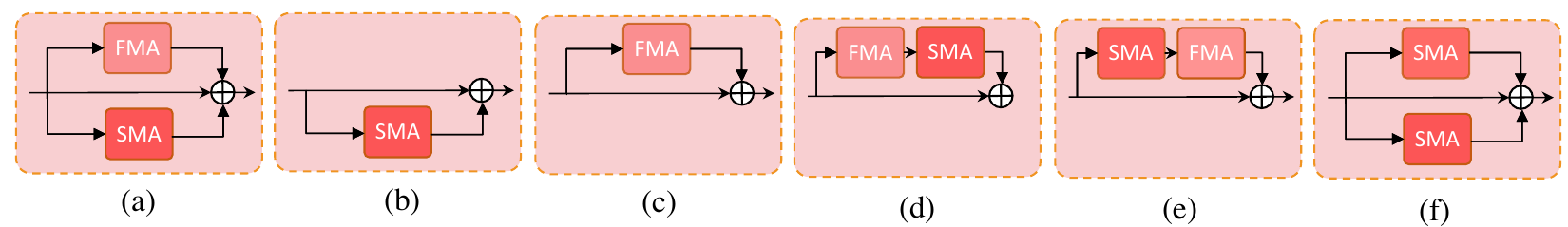}
  \caption{Different variants of SFMA: (a) Proposed SFMA. (b) SMA-only.(c) FMA-only. (d) FMA-SMA-sequential. (e) SMA-FMA-sequential. (f) SMA-SMA-parallel.}
  \label{fig:abla-on-arch}
\end{figure}
\subsection{Ablation study}
We conduct ablation studies to investigate how each design in our SFMA influences the overall performance. Without loss of generality, we use the pre-trained \textit{mbt2018-mean} as the base codec for convenience.

\noindent\textbf{Effect on the architecture of SFMA. }
We first explore the impact of the architecture of SFMA on performance by conducting experiments with various variants of SFMA, as summarized in \cref{fig:abla-on-arch}.  \textbf{(a) Proposed SFMA}: The default configuration of  SFMA.  \textbf{(b) SMA-only}: The FMA branch is removed, retaining only the SMA branch. \textbf{(c) FMA-only}: The SMA branch is removed, retaining only the FMA branch. \textbf{(d) FMA-SMA-sequential}: FMA and SMA are organized sequentially, where SMA is followed by FMA. \textbf{(e) SMA-FMA-sequential}: FMA and SMA are organized sequentially, where FMA is followed by SMA.
\textbf{(f) SMA-SMA-parallel}: two SMAs are configured in parallel.

The results are shown in \cref{tab:abla_on_arch}. We can observe that \textbf{(1)} removing either SMA or FMA leads to performance degradation, and FMA plays a more important role than SMA in the adaptation process, showing the superiority of frequency domain adaptation. \textbf{(2)} parallel configuration outperforms the sequential configuration. \textbf{(3)} barely replicating SMA has more trainable parameters compared to our SFMA, but leads to a performance degradation. This shows that the improvement of our SFMA is not only due to an overall increase in model capacity but rather to its specific structure.

\begin{table}[t]
\centering
\caption{Ablations on different variants of SFMA}
\begin{tabular}{c|cc|cc|cc|c}
\hline
  \multirow{2}{*}{\textbf{Method}} & \multicolumn{2}{c|}{\textbf{Classification}} & \multicolumn{2}{c|}{\textbf{Detection}} & \multicolumn{2}{c|}{\textbf{Segmentation}}& Params  \\
         & BD-rate$\downarrow$ & ~BD-acc$\uparrow$~ & BD-rate$\downarrow$ & BD-mAP$\uparrow$ & BD-rate$\downarrow$ & BD-mAP$\uparrow$ &(M) \\

\hline
 \rowcolor{red!10} 
(a) & -82.00\% & 18.71 & -56.17\%&3.84&-52.65\%&3.17&0.28\\
(b) & -73.82\% & 16.30 & -51.94\%& 3.61& -48.16\%&2.92 &0.16 \\
(c) & -77.40\% & 17.29 & -52.86\%& 3.32& -49.58\%&2.90 &0.12 \\
(d) & -80.49\% & 18.67 & -56.87\%& 3.81& -51.79\% &3.14&0.28 \\
(e) & -83.31\% & 18.50 & -53.77\%& 3.79&  -51.28\% &3.02&0.28 \\
(f) & -79.63\% & 17.59 & -54.91\%& 3.62&  -50.96\%&3.04&0.32 \\
\hline
\end{tabular}
\label{tab:abla_on_arch}
\end{table}

\noindent\textbf{Effect on the position of SFMA.}
We then investigate the impact of the position to plug the SFMA. The results are shown in the Table~\ref{tab:abla_on_position}. From shallow to deep, we first plug SFMA after different stages of encoder $g_a$ and decoder $g_s$. We observe that the SFMA at the shallow layer can benefit more for the rate-accuracy performance, \textit{i.e.,} plugging SFMA after stage1 of $g_a$ and $g_s$ can achieve $69.54\%$ bitrates savings for classification compared to the base codec, while plugging SFMA after stage3 of $g_a$ and $g_s$ can only achieve $64.76\%$ bitrates savings. In addition, employing more SFMA can bring further benefits. 

Then, we observe that SFMAs are indispensable at both the encoder and decoder side, such that removing either would lead to a significant degradation in performance. SFMA in the encoder tends to reduce redundancies in the feature for machine vision tasks.  SFMA in the decoder tends to adapt the feature so that the reconstructed image is more suitable for downstream tasks.

\begin{table}[t]
\centering
\caption{Ablations on the position of SFMA.}
\begin{tabular}{ccc|cc|cc|cc|cc|c}
\hline
\multicolumn{3}{c|}{\textbf{Stage}}& \multirow{2}{*}{\textbf{$g_a$}}& \multirow{2}{*}{\textbf{$g_s$}} & \multicolumn{2}{c|}{\textbf{Classification}} & \multicolumn{2}{c|}{\textbf{Detection}} & \multicolumn{2}{c|}{\textbf{Segmentation}}& Params  \\
    1& 2&3&  && BD-rate$\downarrow$ & ~BD-acc$\uparrow$~ & BD-rate$\downarrow$ & BD-mAP$\uparrow$ & BD-rate$\downarrow$ & BD-mAP$\uparrow$ &(M) \\

\hline
 \checkmark&&& \checkmark& \checkmark&-69.54\% & 15.95 & -50.72\%& 3.39& -48.27\%&2.92 &0.09 \\ 
 & \checkmark&& \checkmark& \checkmark & -69.00\% & 15.69 & -49.96\%& 3.24& -46.40\%&2.82 &0.09 \\
 && \checkmark& \checkmark& \checkmark & -64.76\% & 13.29 & -50.08\%& 3.17& -39.98\%&2.43&0.09  \\
  \rowcolor{red!10} 
  \checkmark& \checkmark& \checkmark& \checkmark& \checkmark & -82.00\%&18.71&-56.17\%&3.84&-52.65\%&3.17&0.28\\\hline
  \checkmark& \checkmark& \checkmark& \checkmark&  & -64.75\% & 13.28 & -38.84\%& 2.88&-38.67\% & 2.28& 0.14\\
    \checkmark& \checkmark& \checkmark& & \checkmark & -61.87\% & 12.40 & -44.05\%& 3.01&-40.43\% & 2.66& 0.14\\
\hline
\end{tabular}
\label{tab:abla_on_position}
\end{table}

\noindent\textbf{Effect on the middle dimension $\hat{C}$.}
\cref{tab:abla_on_dim} reports how middle dimension $\hat{C}$ impacts the rate-accuracy performance and  complexity of SFMA. The results show that enlarging the middle dimension from 1 to 128 can boost performance while leading to a higher training complexity. Thus, we set $\hat{C}$ to 64 to make a better trade-off between performance and complexity.

\begin{table}[b]
\centering
\caption{Ablations on the middle dimension $\hat{C}$ of SFMA.}
\begin{tabular}{c|cc|cc|cc|c}
\hline
  \multirow{2}{*}{\textbf{$\hat{C}$}} & \multicolumn{2}{c|}{\textbf{Classification}} & \multicolumn{2}{c|}{\textbf{Detection}} & \multicolumn{2}{c|}{\textbf{Segmentation}}& Params  \\
         & BD-rate$\downarrow$ & ~BD-acc$\uparrow$~ & BD-rate$\downarrow$ & BD-mAP$\uparrow$ & BD-rate$\downarrow$ & BD-mAP$\uparrow$ &(M) \\

\hline
1&-60.41\%&12.58&-48.58\% &2.93 & -45.24\%& 2.40& 0.005\\
32&-78.74\%&17.52&-53.61\% &3.53 & -48.50\%& 2.92& 0.14\\
 \rowcolor{red!10} 
64 & -82.00\% & 18.71 &-56.17\%&3.84&-52.65\%&3.17&0.28\\
128 & -83.61\% & 19.50 &-63.28\% &4.11 &-60.50\% &3.76 & 0.62\\
\hline
\end{tabular}
\label{tab:abla_on_dim}
\end{table}

\begin{figure*}[h]
\begin{center}
\subfloat{\includegraphics[width=1\linewidth]{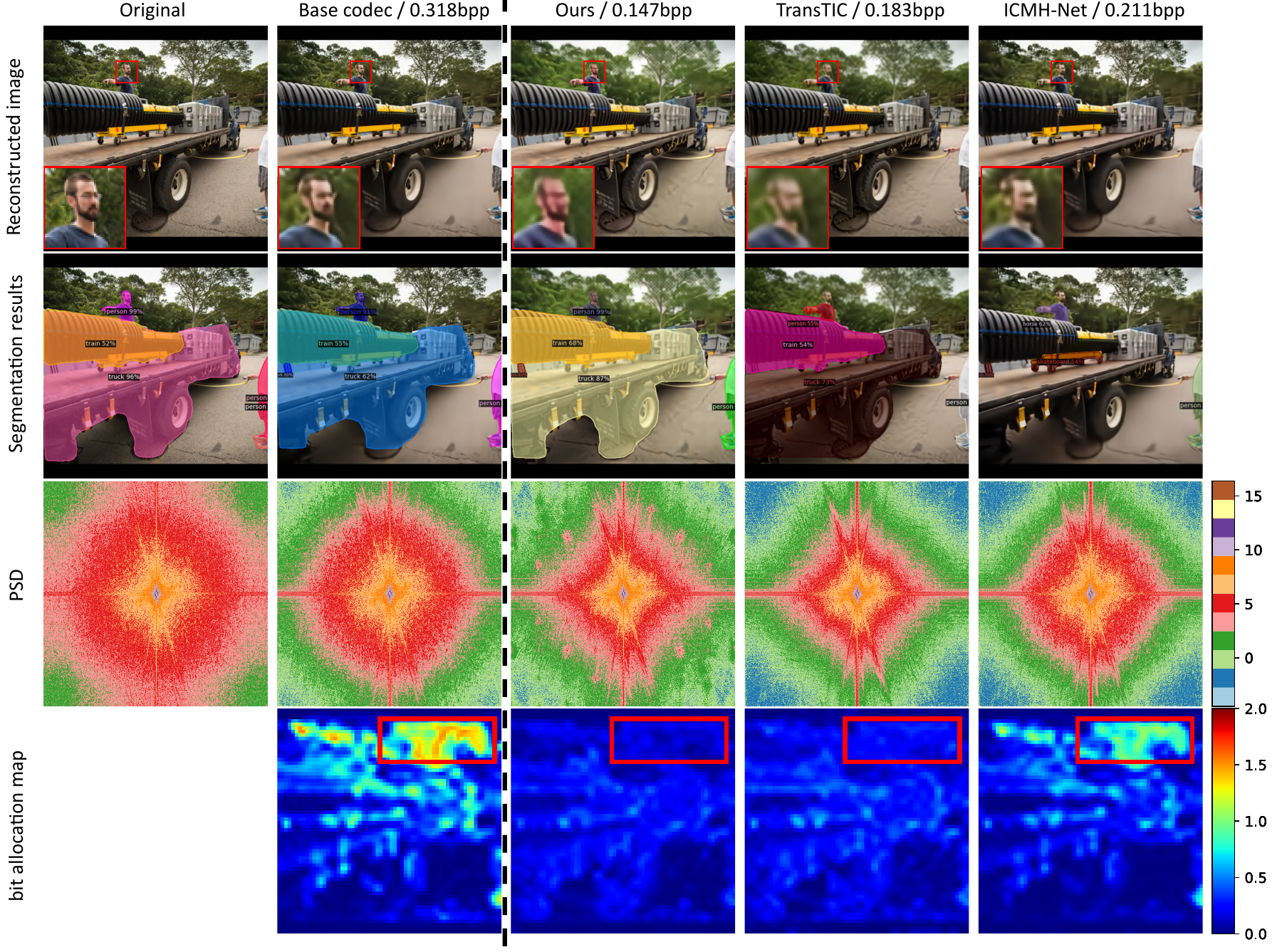}}
\caption{Qualitative comparison of our Adapt-ICMH with other ICMH methods. \textbf{First row:} The original image and decoded image of each method. We show the decoded images for machine vision of three ICMH methods (left). \textbf{Second row:} The instance segmentation results of each image. \textbf{Third row:} The log power spectral density maps of each image. \textbf{Bottom row:} The bit allocation maps for $\bm{\hat{y}}$ of each method.  }
\label{fig:quantitative-results}
\end{center}
\end{figure*}
\subsection{Qualitative Results}
\cref{fig:quantitative-results} shows the qualitative results of our Adapt-ICMH and other methods for instance segmentation. Our approach achieves superior segmentation results with a lower bitrate (0.147bpp). In comparison to other methods, Adapt-ICMH effectively minimizes the bitrates of the background while enhancing the high-frequency edge details of the semantic object.
We also observe that our method's decoded image for machine vision exhibits some distortion in terms of overall chromaticity and brightness. However, these low-frequency distortions do not affect the segmentation result, which supports the goal of our SFMA. 

\subsection{Towards scalable coding for machine and human vision.} 
Scalable coding for machine and human vision is of practical significance to adapt the transmission among diverse demands.
The framework of Fig~\ref{fig:overview}, followed by~\cite{chen2023transtic,liu2022improving}, is designed to produce a single bitstream tailored for each individual task.  However, our proposed SFMA can also benefit existing scalable coding methods to achieve better scalable coding performance for machine and human vision. To demonstrate it, we integrate our SFMA and the spatial-channel mask generator proposed by \textit{ICMH-Net}~\cite{liu2023icmh} with the base codec. Specifically, we only plug our SFMA into the decoder $g_s$ and utilize spatial-channel mask generator to produce two bitstreams, the truncated one is decoded by the $g_s$ with SFMAs to reconstruct image for machine vision task, while the full bitstream is decoded by the $g_s$ without SFMAs to reconstruct image for human vision. See Appendix F for detailed network architecture for scalable coding.

\begin{figure}[t]
  \centering
  \includegraphics[width=1\linewidth]{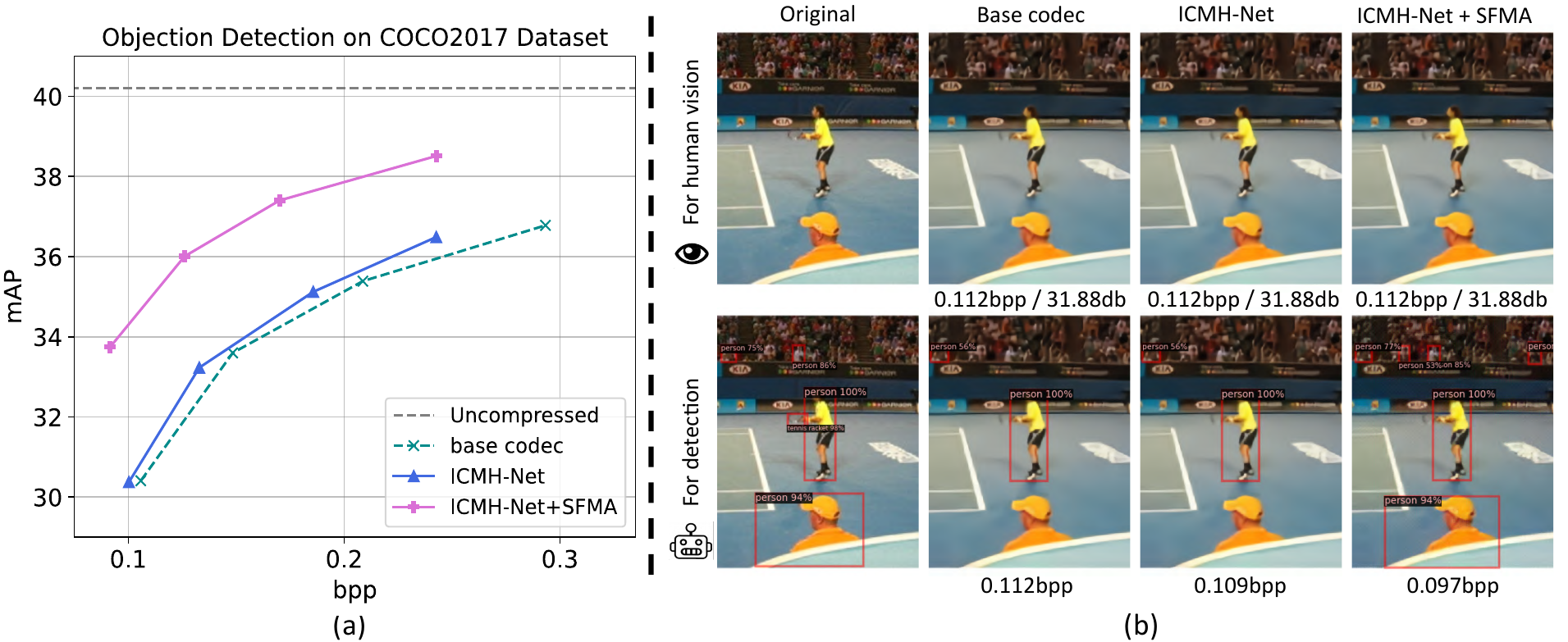}
  \caption{Quantitative and qualitative comparison of each method for the object detection task. Pre-trained \textit{mbt2018-mean}~\cite{minnen2018joint} is used as the base codec. We present the bpp/PSNR results for the decoded image for human vision and the bpp results for the decoded image for the detection task.}
  \label{fig:scalable_results}
\end{figure}

\cref{fig:scalable_results} show the comparison of rate-accuracy performance and qualitative results. It demonstrates the superior performance of our SFMA, which significantly improves the rate-accuracy performance of \textit{ICMH-Net}. Note that 
 we do not alter the base codec, so the visual quality of the image reconstructed from the full bitstream is the same as the base codec.
\section{Conclusion}

 In this paper, we develop a novel adapter-based tuning framework for ICMH, named Adapt-ICMH. Specifically, we fine-tune the pre-trained image codec optimized for human vision on multiple machine vision tasks, which is achieved by the proposed spatial-frequency modulation adapter (SFMA). Our SFMA can efficiently reduce the latent redundancies in both spatial and frequency domains, and effectively adapt the decoded image for downstream machine vision tasks. Our method is plug-and-play and compatible with all existing learned image compression models. Experiments show that our method consistently outperforms other ICMH methods in various machine vision tasks, even with a reduced number of trainable parameters and computational complexity. In the future work, we will extend our work to video coding for machines and humans.
 
\section{Acknowledgement}
This work was supported in part by the National Natural Science Foundation of China under Grant 62125109, Grant 61931023, Grant 61932022, Grant 62371288, Grant 62320106003, Grant 62301299, Grant T2122024, Grant 62120106007, China Postdoctoral Innovation Talents Support Program (BX20230184).

\appendix

\section{Preliminary of Learned Image Compression}
\label{sectionA}
Learned image compression model typically consists of two core modules: nonlinear transform and entropy model.  The nonlinear transform including analysis transform (\textit{i.e.,} encoder, $g_a$) and synthesis transform (\textit{i.e.,} decoder, $g_s$). Given the raw image $\bm{x}$, the encoder $g_a(\cdot)$ maps it to the latent representation $\bm{y}$. Then, quantization operator $Q(\cdot)$ discretizes $\bm{y}$ to $\bm{\hat{y}}$. Finally, the reconstructed image $\bm{\hat{x}}$ is obtained by feeding the quantized latent $\bm{\hat{y}}$ to the synthesis transform $g_s(\cdot)$.
This process can be formulated as follows:
\begin{align}
    \bm{y}&= g_a(\bm{x};\bm{\theta_{a}}),\\
    \bm{\hat{y}}&=Q(\bm{y}),\\
    \bm{\hat{x}}&=g_s(\bm{\hat{y}};\bm{\theta_{s}}),
\end{align}
where $\bm{\theta_{a}}$ and $\bm{\theta_{s}}$ are the trainable parameters of encoder and decoder, respectively. To encode $\bm{\hat{y}}$ losslessly, the entropy model with side information $\bm{\hat{z}}$ is used to model each element of $\bm{\hat{y}}$ as a Gaussian distribution with the predicted parameters of mean $\bm{\mu}$ and scale $\bm{\sigma}$:
\begin{align}
    p_{\bm{\hat{y}}|\bm{\hat{z}}}(\bm{\hat{y}}|\bm{\hat{z}}) = \mathcal{N}(\bm{\mu},\bm{\sigma}^2),
\end{align}
Then, we can calculate the bitrates by:
\begin{align}
    R= -\log_2 p_{\bm{\hat{y}}|\bm{\hat{z}}}(\bm{\hat{y}}|\bm{\hat{z}}) -\log_2 p_{\bm{\hat{z}}}(\bm{\hat{z}}).
\end{align}

If the decoded image $\bm{\hat{x}}$ is reconstructed for human vision, we usually measure its visual quality by calculating the mean square error over the raw image. However, this metric is not suitable when the image is  reconstructed for machine vision. In this paper, we adopt the perceptual distortion loss to optimize the adpaters for better task performance.  

\section{More Discussion about ICMH}
\label{sectionB}
We further discuss the differences and advantages of our approach over existing methods across various frameworks in detail.  
\subsubsection{Scalable coding framework}
Earlier work~\cite{choi2022scalable,liu2021semantics,yang2021towards,liu2023icmh} achieves scalable coding for human and machine vision. Choi~\emph{et al.}~\cite{choi2022scalable} adopts a mutli-task pipeline that needs to jointly train the whole codec (\textit{i.e.,} one encoder and multiple task-specific decoders) from scratch. Despite its superior performance on machine vision tasks, the multi-task pipeline dramatically degrades the R-D performance on human vision (about 1dB degradation in PSNR compared to optimizing only for human vision). In addition, training from scratch also brings huge training overhead and is restricted in adapting to newly coming machine vision tasks. 
In contrast, our Adapt-ICMH efficiently fine-tune the pre-trained human vision-oriented image codecs on machine vision without compromising the R-D performance of the original codec.

ICMH-Net~\cite{liu2023icmh}, also based on the pre-trained image codec, selects partial quantized latent as the base bitstream for specific machine vision task.  It does not sacrifice the R-D performance since the pre-trained image codec is frozen. However, the machine vision performance is unsatisfactory due to the use of a binary mask generated by Gumbel-Softmax to filter redundant latent, which is suboptimal and difficult to optimize stably. In contrast, our SFMA achieves latent adaptation through spatial-frequency modulation rather than simple spatial selection. Additionally, we have demonstrated the importance of updating shallow latent, while ICMH-Net only adapts the deepest latent.

Although our original framework  wasn't specifically tailored for scalable coding, our SFMA can still integrate with scalable coding frameworks (\emph{e.g.,} ICMH-Net) to significantly improve their rate-accuracy performance.

\subsubsection{Single bitstream framework.}

\cite{chen2023transtic,liu2022improving,feng2023semantically,fischer2022boosting} produce single bitstreams for each individual task and  our framework also belongs to this category.  Feng~\emph{et al.}~\cite{feng2023semantically} uses the group mask generated by the pre-analysis recognition models (\emph{e.g.,} detection and segmentation models) to implement ROI coding for machine tasks. But it's not practical to deploy so many pre-analysis models at encoder side (usually user side). Liu~\emph{et al.}~\cite{liu2022improving} 
performs channel selection and produces individual bitstreams for each specific machine task, but it also requires task-specific decoder.
Chen~\emph{et al.}~\cite{chen2023transtic} transfers the transformer-based image codec from human vision to machine vision by visual prompt tuning, but it fails to be compatible with CNN-based image codecs, and brings significant additional computational complexity. In contrast, our proposed  SFMA is lightweight and compatible with almost all the existing LIC models with the original image codec shared across human and machine vision.

\subsubsection{generalized bitstream framework.}
There are also works~\cite{feng2022image,bai2022towards,feng2023prompt} that uses a single bitstreams for multiple tasks. Feng~\emph{et al.}~\cite{feng2022image} learns a compressed omnipotent feature for multiple machine vision tasks. Bai~\emph{et al.}~\cite{bai2022towards}  jointly optimizes the quantized latent for human perception and classification task. However, it's still challenge to trade-off between multiple tasks and find a optimal solution.

\section{Details of Training Setting}
\label{sectionC}

\subsubsection{Task-specific Perceptual Distortion Loss.}

In Eq.(1) of our main paper, we use the task-specific perceptual distortion loss $\mathcal{D}$ to optimize our SFMAs for machine vision tasks, allowing us to train the task-specific module without accessing the task-related label.

\begin{wrapfigure}[14]{r}{0.45\textwidth}
\vspace{-15pt}
  \centering
  \includegraphics[width=0.99\linewidth]{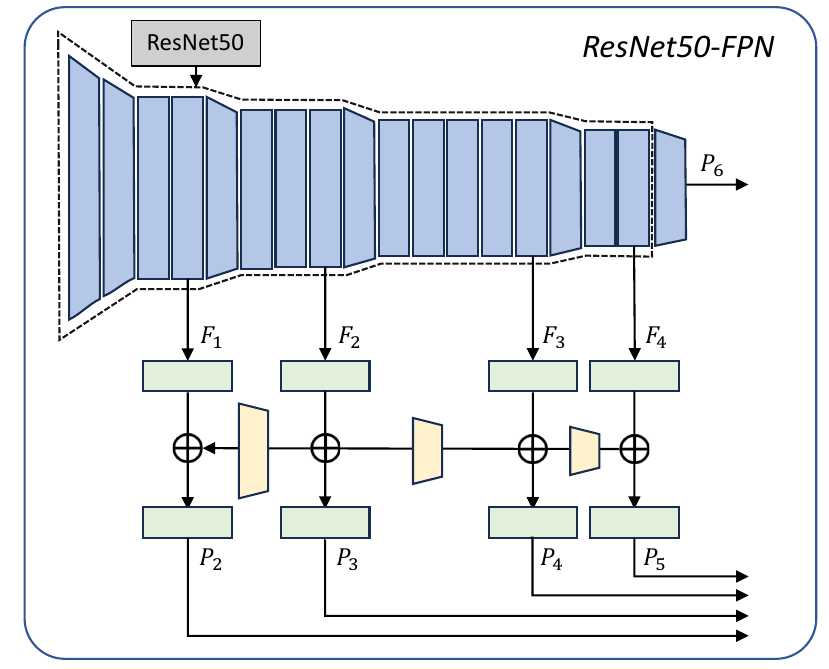}
  \caption{Network architecture of ResNet50-FPN. }
  \label{fig:appendix:resnet50fpn}
\end{wrapfigure}

Specifically, we follow~\cite{chen2023transtic} to use pre-trained ResNet50\footnote{\url{https://download.pytorch.org/models/resnet50-0676ba61.pth}}~\cite{he2016deep}, Faster 
RCNN\footnote{\url{https://dl.fbaipublicfiles.com/detectron2/COCO-Detection/faster\_rcnn\_R\_50\_FPN\_3x/137849458/model\_final\_280758.pkl}}~\cite{ren2015faster}, and Mask
RCNN\footnote{\url{https://dl.fbaipublicfiles.com/detectron2/COCO-InstanceSegmentation/mask_rcnn_R_50_FPN_3x/137849600/model_final_f10217.pkl}}~\cite{he2017mask}
as the off-the-shelf recognition model for classification, object detection, and instance segmentation, respectively. \cref{fig:appendix:resnet50fpn} shows the network architectures of ResNet50-based Feature Pyramid Network (FPN), which is the backbone network of Faster RCNN~\cite{ren2015faster} and Mask RCNN~\cite{he2017mask}.
In particular,  we take the features output by ResNet50 (\textit{i.e.,} $F_1, F_2, F_3,$ and $F_4$) to evaluate the perceptual loss for classification task:
\vspace{-5pt}
\begin{align}
\mathcal{D}(\bm{x}, \hat{\bm{x}}, \mathcal{G}) = \frac{1}{4}\sum_{j=1}^{4}\operatorname{MSE}\left(F_j(\bm{x}),F_j(\bm{\hat{\bm{x}}})\right),
\end{align}
where $\bm{x}$ and $\hat{\bm{x}}$ denotes the raw and reconstructed images, respectively.  For object detection and instance segmentation tasks, we take the features output by FPN (\textit{i.e.,} $P_2, P_3, P_4, P_5,$ and $ P_6$) to calculate the perceptual loss:
\begin{align}
\mathcal{D}(\bm{x}, \hat{\bm{x}}, \mathcal{G}) = \frac{1}{5}\sum_{j=2}^{6}\operatorname{MSE}\left(P_j(\bm{x}),P_j(\bm{\hat{\bm{x}}})\right).
\end{align}
\subsubsection{Hyperparamters of Training}
\cref{tab:appendix:hyperparameters} details the training hyperparamters  of our all experiments for different machine tasks. 
We use NVIDIA GeForce RTX 4090 and Intel Xeon Platinum 8260 to conduct all our experiments.

\begin{table}[]
\vspace{-10pt}
\caption{Training hyperparamters for experiments in the main paper. }
    \centering
   \resizebox{0.98\textwidth}{!}{ \begin{tabular}{c|c|c|c}
    \hline
         & Classification & Detection & Segmentation \\ 
         \hline
        Optimizer & Adam & Adam &Adam \\
        Batch size  &16 &8 &8\\
        Trade-off term $\lambda $ & ~[1.8, 3.5, 6.7, 13]~ & ~[5, 10, 20, 50]~ & ~[5, 10, 20, 50]~\\
       Epochs    & 5& 40&40 \\ 
       Learning rate schedule & MultiStepLR&- & -\\
       Milestones & [2, 4] & -& -\\
       Learning rate decay  & 0.5& -& -\\
       Base learning rate& 1e-4& 1e-4& 1e-4\\ \hline    
    \end{tabular}}
    \label{tab:appendix:hyperparameters}
    \vspace{-10pt}
\end{table}
\clearpage
\section{Details of Reproduction for Other Methods}
\label{sectionD}
\begin{figure*}[htp]
\begin{center}
\subfloat[]{\includegraphics[width=.45\linewidth]{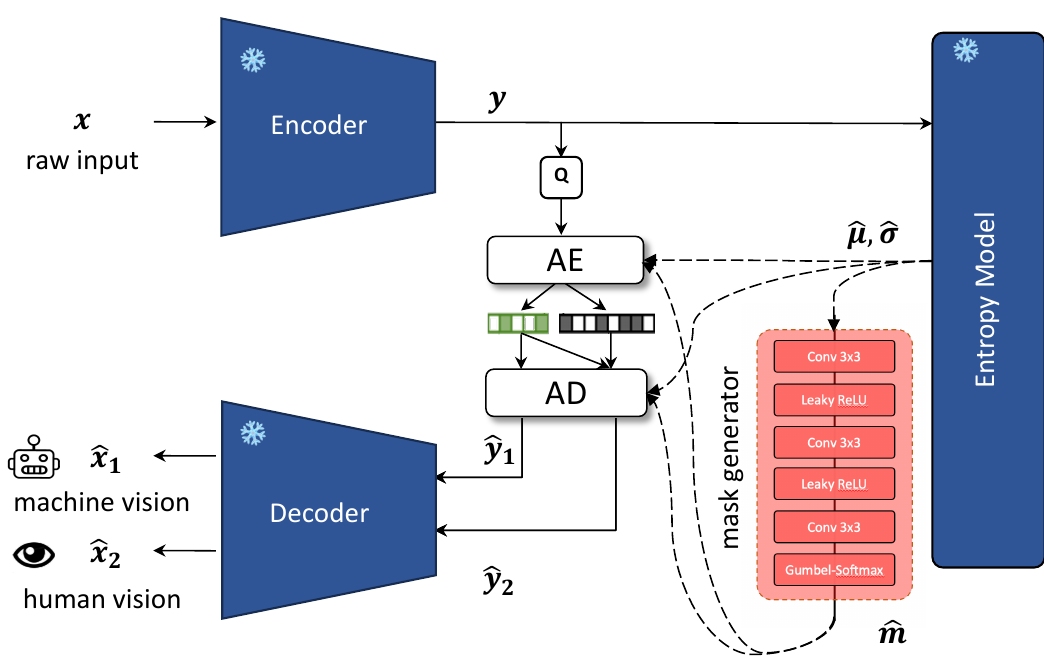}} \hspace{0.7cm}
\subfloat[]{\includegraphics[width=.45\linewidth]{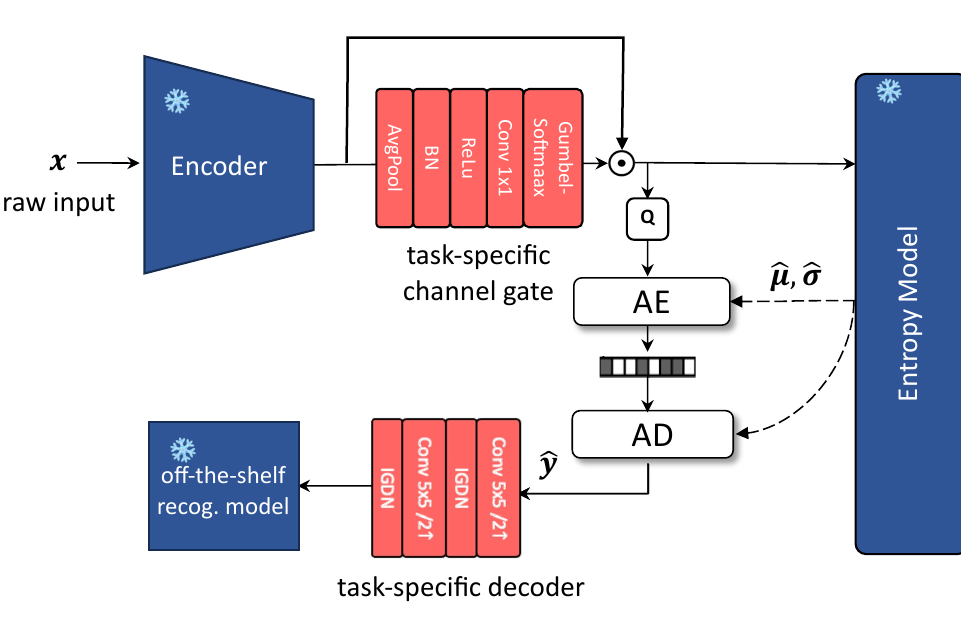}}
\caption{Architecture of ICMH-Net~\cite{liu2023icmh} (left) and channel selection~\cite{liu2022improving} (right). }
\label{fig:appendix-reproduce}
\end{center}
\end{figure*}

We have compared our framework with SOTA tuning-based methods~\cite{chen2023transtic,liu2023icmh,liu2022improving}. We use the result of TransTIC~\cite{chen2023transtic} published in the paper. However, \cite{liu2023icmh} and~\cite{liu2022improving} used different pre-trained recognition models from us\footnote{~\cite{liu2022improving} used the Deep Lab V3 for segmentation task. Although ~\cite{liu2023icmh} claimed that they used ResNet50 for classification and Faster RCNN for detection, they didn't disclose which pre-trained checkpoints they utilized.}, we cannot directly compare our framework with theirs. Since \cite{liu2023icmh} and~\cite{liu2022improving} are not open source, we reproduce their results by using the same training, evaluation settings, and dataset as ours. \cref{fig:appendix-reproduce} shows the their sketch of architecture, we set the temperature parameter of hard version of Gumbel-Softmax  to 1  for~\cite{liu2023icmh,liu2022improving}.

\section{Details of Architecture for Different Base Codecs}
\label{sectionE}
\begin{figure*}[htp]
\begin{center}
\subfloat[\textit{Lu2022-TIC} model~\cite{lu2022transformer} as base codec. STB denotes the Swin-Transformer Block~\cite{liu2021swin}. Following~\cite{chen2023transtic}, we adopt the simplified version of~\cite{lu2022transformer} for fair comparison. ]{\includegraphics[width=.8\linewidth]{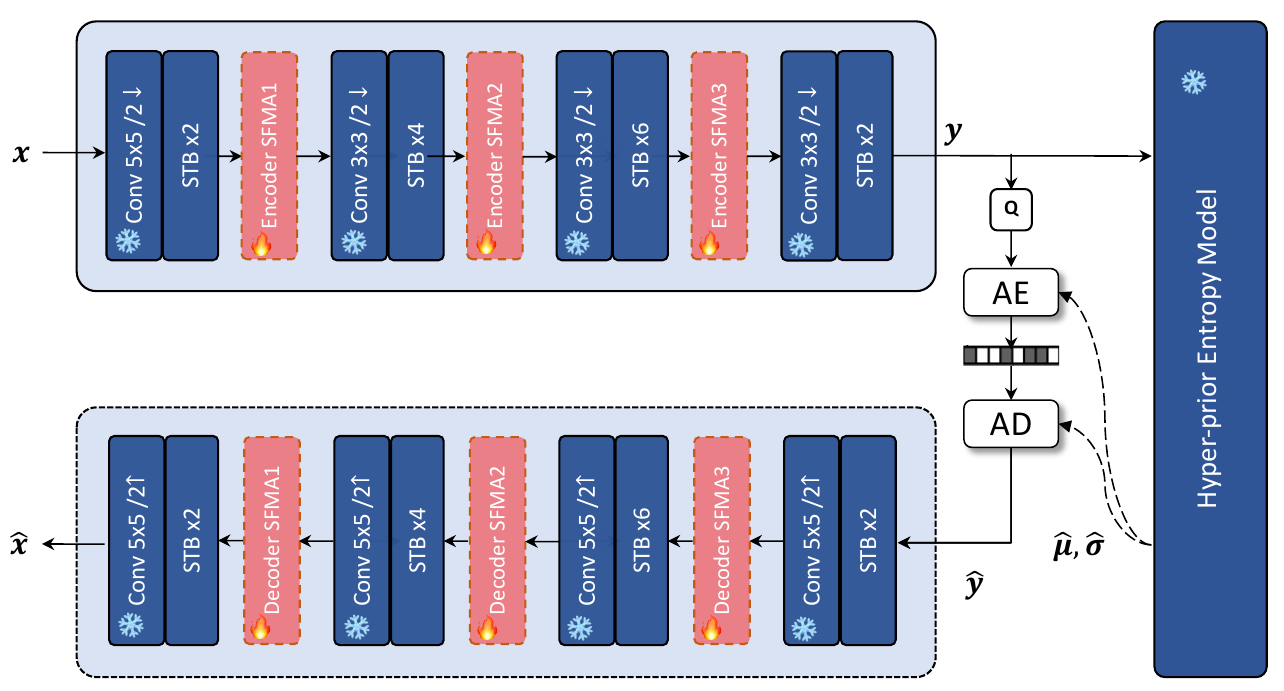}}\\
\subfloat[\textit{Cheng2020-anchor} model~\cite{cheng2020learned} as base codec. ResBlock denotes the residual block, $\backslash2\downarrow$ denotes a stride on the first convolution, and $\backslash2\uparrow$ denotes a sub-pixel upsampling on the last convolution. ]{\includegraphics[width=.8\linewidth]{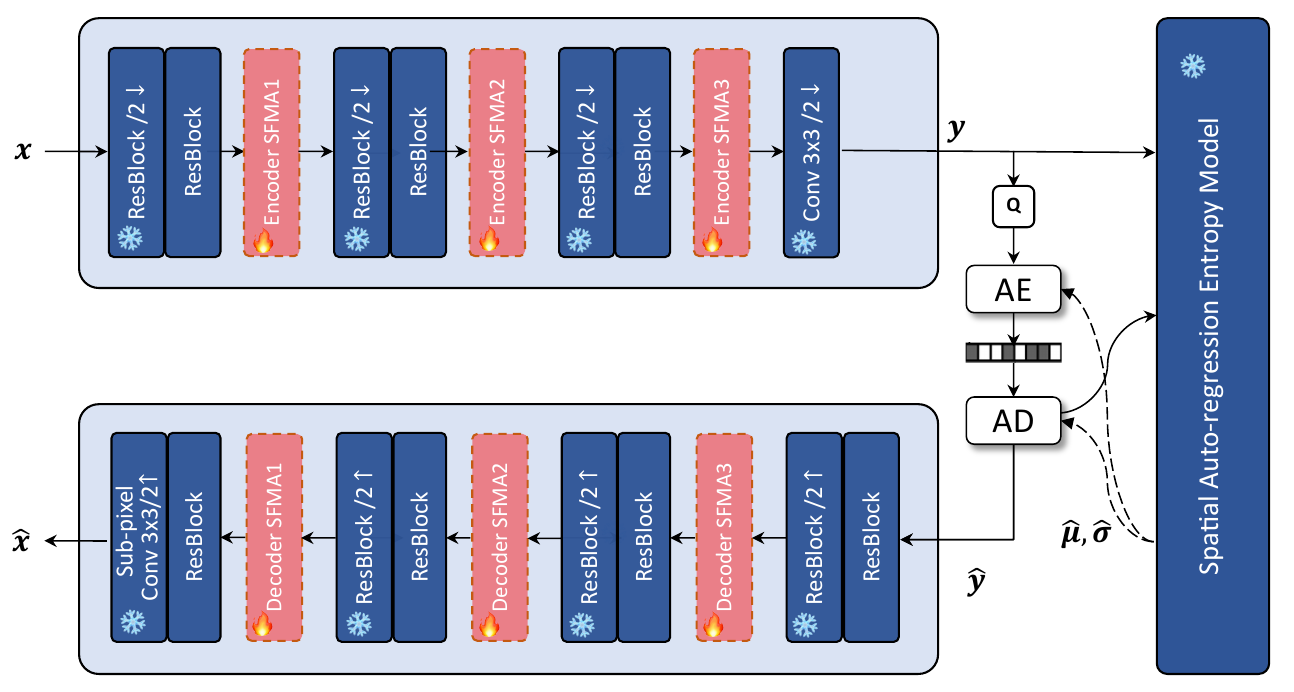}}\\
\subfloat[\textit{mbt2018-mean} model~\cite{minnen2018joint} as base codec. GDN denotes the Generalized Divisive Normalization layer~\cite{balle2015density}.]
{\includegraphics[width=.8\linewidth]{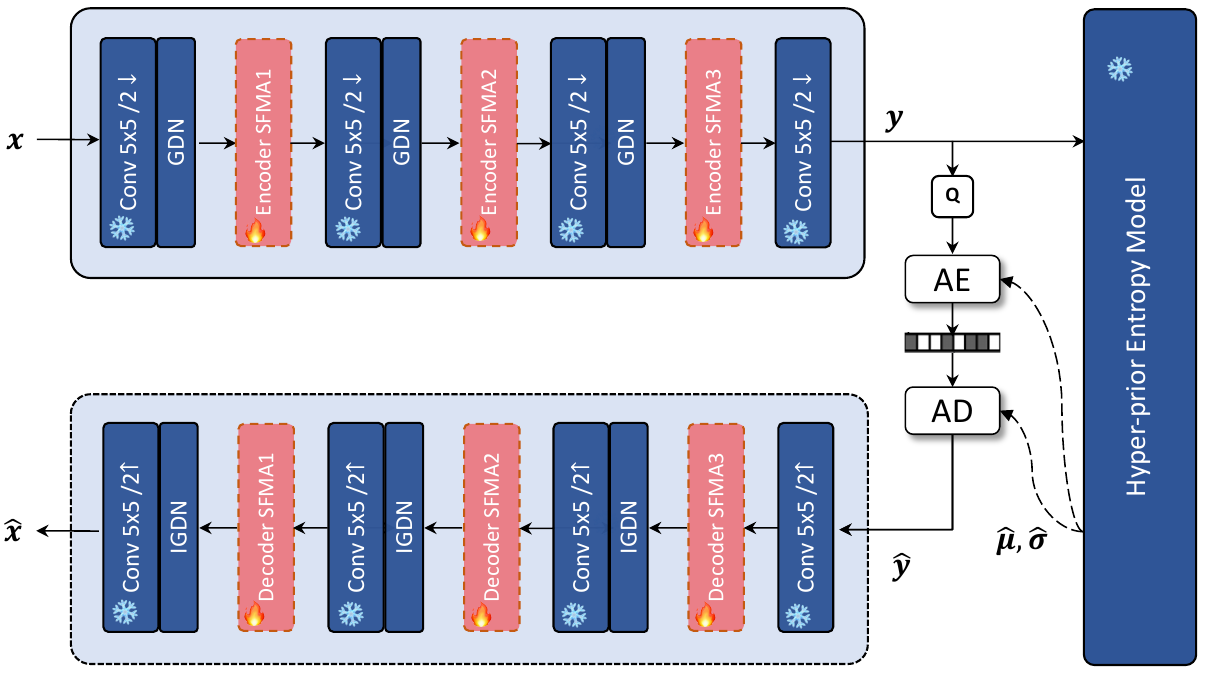}}
\caption{Details of network architecture for different base codecs.}
\label{fig:appendix-detailed-architecture}
\end{center}
\end{figure*}

We illustrate the detailed network architecture for our Adapt-ICMH framework incorporating different image codec methods in \cref{fig:appendix-detailed-architecture}. Note that these image codecs have different nonlinear transforms and entropy models, but our SFMA consistently shows the ability to achieve superior rate-accuracy performance for machine vision tasks, and do not affect the visual quality for human vision of the base codec.
We also include the source of the pre-trained checkpoint of base codecs in the \cref{tab:appendix:source}.

\begin{table}[htbp]
    \centering
    \caption{Source path of the implementations and pre-trained weights of base codecs}
    \begin{tabular}{c|c}  \hline
         Model &  Source Path \\ \hline
         \textit{Lu2022-TIC}&  \href{https://github.com/NYCU-MAPL/TransTIC/tree/master}{https://github.com/NYCU-MAPL/TransTIC/tree/master} \\
        \textit{Cheng2020-anchor}& \href{https://github.com/InterDigitalInc/CompressAI/tree/master}{https://github.com/InterDigitalInc/CompressAI/tree/master} \\
       \textit{mbt2018-mean}& \href{https://github.com/InterDigitalInc/CompressAI/tree/master}{https://github.com/InterDigitalInc/CompressAI/tree/master}  \\ \hline
    \end{tabular}
    \label{tab:appendix:source}
\end{table}
\section{Details of Scalable Coding Pipeline}
\label{sectionF}
\begin{figure}[h]
  \centering
  \includegraphics[width=0.8\linewidth]{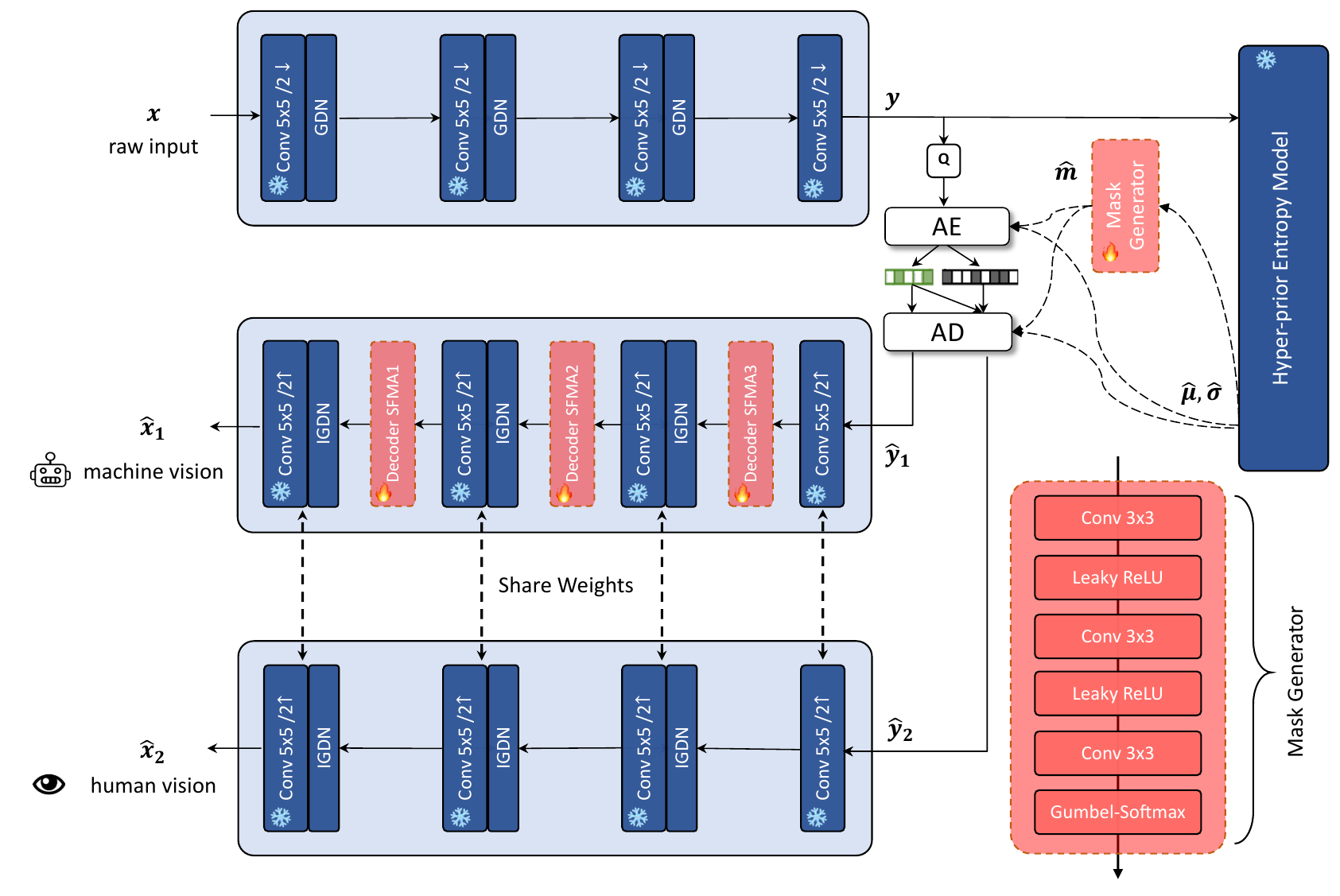}
  \caption{Our scalable coding framework. The mask generator is proposed by~\cite{liu2023icmh} and we adopt pre-trained \textit{mbt2018-mean} as base codec. }
  \label{fig:appendix:scalable}
\end{figure}
\begin{figure*}[htp]
\begin{center}
\subfloat[]{\includegraphics[width=.4\linewidth]{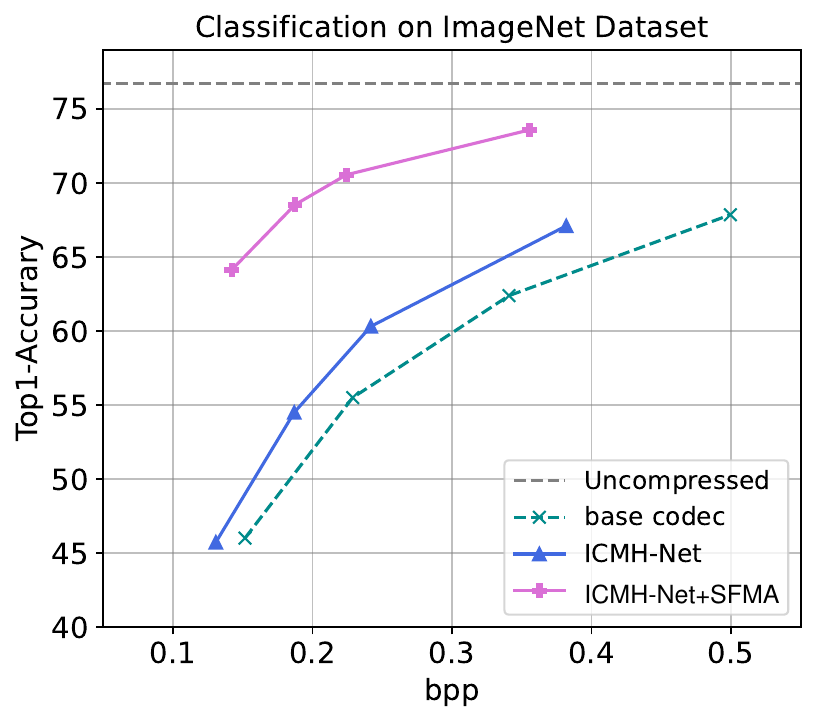}} \hspace{0.5cm}
\subfloat[]{\includegraphics[width=.4\linewidth]{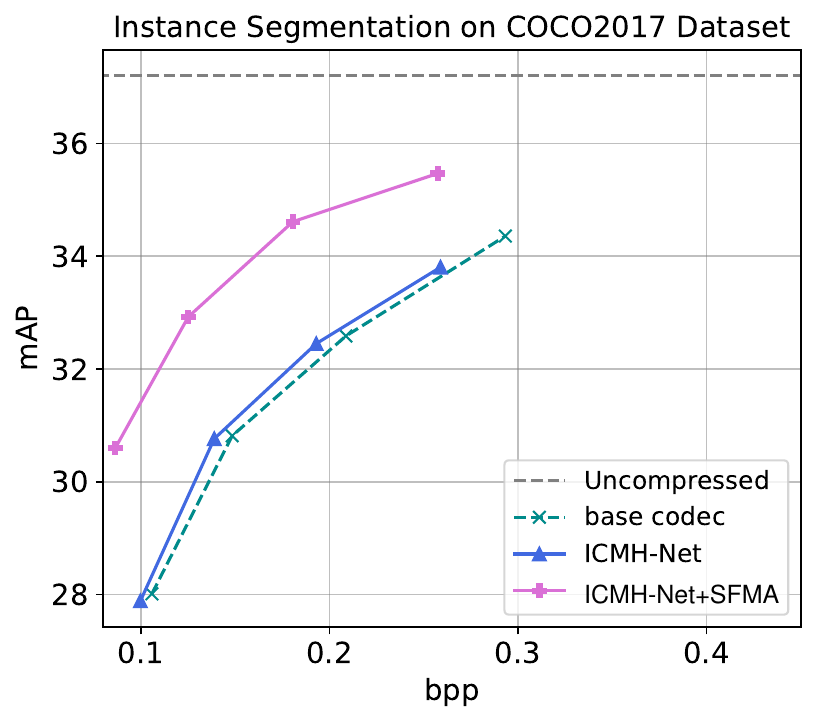}}
\caption{Rate-accuracy results of classification  (left) and instance segmentation   (right).  }
\label{fig:appendix-scalable-results}
\end{center}
\end{figure*}

We present the framework of scalable coding for machine and human vision in \cref{fig:appendix:scalable}. Specifically, we only inject  the SFMAs in the decoder of base codec and adopt the mask generator~\cite{liu2023icmh} to select the latent to be transmitted for machine vision. Specifically, the binary spatial-channel mask $\hat{\bm{m}}$ is derived by  inputting the entropy parameters $\hat{\bm{\mu}}$ and $\hat{\bm{\sigma}}$ into the mask generator.  Thus, we obtain the masked quantized latent $\hat{\bm{y}}_1$ (\textit{i.e., }$\hat{\bm{y}}_1 = \hat{\bm{m}}\cdot \operatorname{Q}(\hat{\bm{y}})$) which is encoded as the base layer, while the remaining latent is encoded as the enhanced layer.  On the decoder side, the base layer $\hat{\bm{y}}_1$ is decoded by the decoder with SFMAs to obtain the reconstructed image $\hat{\bm{x}}_1$ for machine vision, while the full latent $\hat{\bm{y}}_2$ is decoded by the decoder without SFMAs to obtain the reconstructed image $\hat{\bm{x}}_2$ for human vision. 

We provide the additional rate-accuracy results on classification and instance segmentation tasks in \cref{fig:appendix-scalable-results}. It demonstrates that our proposed SFMA can significantly benefit existing scalable coding framework, \emph{i.e.,} ICMH-Net~\cite{liu2023icmh}.

\section{Analysis on the Computational Complexity}
\label{sectionG}
We compare the computational complexity of our framework with others. \cref{tab:appendix:flops} shows that our framework only introduces small computational complexity and the increase on the latency time can be ignored. Although full fine-tuning can achieve a satisfactory rate-accuracy performance without an increased computational complexity, it requires to store and deploy a copy of the entire codec. Channel selection~\cite{liu2022improving} can reduce the complexity at the decoder side, since it uses a lightweight task-specific decoder (shown in \cref{fig:appendix-reproduce}) for each machine vision task. However, it cannot take advantage of the powerful knowledge of the pre-trained synthesis transform, resulting in degraded rate-accuracy performance in the complex vision tasks (\emph{e.g.,} detection and segmentation) compared to the base codec.

\begin{table}[h]
\renewcommand{\arraystretch}{1.2}
\centering
\caption{Comparison on computational complexity evaluated on the ImageNet-val~\cite{deng2009imagenet} dataset for classification task. We do not include the computation on the off-the-shelf recognition model. Ours-$n$ indicates $n$ middle dimensions. The BD-acc is presented for rate-accuracy performance comparison with the base codec of \emph{Lu2022-TIC}~\cite{lu2022transformer} as the anchor. }
\begin{tabular}{lcccccc}
\hline
\multirow{2}{*}{\textbf{Model}}& \multicolumn{2}{c}{\textbf{KMACs/pixel}~~~~} & \multicolumn{2}{c}{\textbf{Latency (ms)}} & \textbf{\#Trainable}  & \multirow{2}{*}{\textbf{BD-acc $\uparrow$}}\\
\cline{2-5}
        & \textbf{Enc.}          & \textbf{Dec.}          & \textbf{Enc.}          & \textbf{Dec.}             &     \textbf{ Params(M)}     &      \\ \hline
 base codec &142.5&188.5&120.1&35.8&-&-\\ \hline
full fine-tuning &142.5&188.5&120.1&35.8&7.51&\textbf{17.6\%}\\
ICMH-Net~\cite{liu2023icmh} &159.1&205.1&126.7&43.3&3.98&3.3\%\\
channel selection~\cite{liu2022improving} &142.6&25.1&121.1&10.1&0.91&6.2\%\\
TransTIC~\cite{chen2023transtic} &332.0&202.6&146.2&40.2&1.61&9.9\%\\\hline
Ours-32 &149.7&195.7&121.7&37.6&0.14&16.4\%\\ 
Ours-64 &157.2&203.2&123.2&40.1&0.28&16.9\%\\ 
Ours-128 &173.4&219.4&124.1&40.9&0.62&\textbf{17.6\%}\\ \hline
\end{tabular}
\label{tab:appendix:flops}
\end{table}

\section{Application on Larger Transformer-based Image Codecs}
\label{sectionH}
We further demonstrate the effectiveness of our framework on larger scale transformer-based image codecs, including \emph{STF}~\cite{zou2022devil}, \emph{TCM}~\cite{liu2023learned}, and \emph{FAT}~\cite{li2023frequency}. We conduct performance comparison with other ICMH frameworks~\cite{chen2023transtic,liu2022improving}, it is noted that  \textit{ICMH-Net}~\cite{liu2023icmh} cannot support the channel-wise autoregression entropy model used in~\cite{zou2022devil,liu2023learned,li2023frequency}. Specifically, we train each methods for one rate-accuracy point and report the classification results in \cref{tab:more_transformer}.  We observe that ours method still outperform other methods with seldom trainable parameters (less than 1\% of the base codec).

\begin{table}
\centering
\caption{Classification comparison on ImageNet-\textit{val} dataset using more large-scale transformer-based image codecs, including  \emph{STF}~\cite{zou2022devil}, \emph{TCM}~\cite{liu2023learned}, and \emph{FLIC}~\cite{li2023frequency}. Acc. denotes the top-1 accuracy. }
\begin{tabular}{l|rr|r}
\hline
  \multirow{2}{*}{\textbf{Method}} & \multicolumn{2}{c|}{\textbf{Classification}} & \textbf{Trainable}\\
         & bpp$\downarrow$ & ~Acc. $\uparrow$~  & Params $\downarrow$(M) \\
\hline
\multicolumn{4}{c}{\textit{STF}~\cite{zou2022devil}} \\
\hline
 \rowcolor{gray!20} 
 full fine-tuning & \textbf{0.2559}&\textbf{75.98}&99.85(100\%)\\
TransTIC~\cite{chen2023transtic}  &0.4910&74.51 &1.15(1.2\%)\\
 channel selection~\cite{liu2022improving}  & 0.5217 &73.21 &2.49(2.5\%)\\
 Ours  &\underline{0.3418} &\underline{75.18}&0.26(0.3\%)\\
\hline
\multicolumn{4}{c}{\textit{TCM}~\cite{liu2023learned}} \\
\hline
 \rowcolor{gray!20} 
 full fine-tuning  &\textbf{0.2494}&\underline{75.83}&76.56(100\%)\\
  TransTIC~\cite{chen2023transtic}  &0.4252&75.60  &1.30(1.7\%)\\
  channel selection~\cite{liu2022improving}  &0.4703& 73.37 &2.73(3.6\%)\\
 Ours  &\underline{0.3273}&\textbf{75.87} &0.53(0.7\%)\\
\hline
\multicolumn{4}{c}{\textit{FLIC}~\cite{li2023frequency}}\\
\hline
 \rowcolor{gray!20} 
 full fine-tuning & \textbf{0.2403}&\textbf{75.93}&70.97(100\%)\\
TransTIC~\cite{chen2023transtic}  &0.3775 &75.65  &1.22(1.7\%)\\
channel selection~\cite{liu2022improving} & 0.4067&73.46 &2.73(3.8\%)\\
 Ours  &\underline{0.3274}&\underline{75.72} &0.36(0.5\%)\\
\hline
\end{tabular}
\label{tab:more_transformer}
\end{table}
\section{More Ablation Studies}
\label{sectionI}
\subsection{Plugging SFMAs into Entropy Model}
Our proposed SFMA is designed to fine-tune the nonlinear transform of the base codec for machine vision task. In this section, we also plug SFMAs into the entropy model to further explore its effectiveness. Specifically, SFMAs are plugged into the intermediate layer of hyper encoder $h_a$ and hyper decoder $h_s$, which is similar to the process of SFMA for $g_a$ and $g_s$. From \cref{tab:appendix-on-entropy-model} we observe that further plugging SFMAs into the entropy model cannot bring significant performance improvement, but introduces more model complexity. Thus, we decide to only plug SFMAs into the nonlinear transform. This also demonstrates that it's the nonlinear transform rather than the entropy model that is the key difference between human and machine vision-oriented image compression.

\begin{table}[htp]
\vspace{-10pt}
\centering
\caption{Ablations on plugging SFMAs into entropy model}
\begin{tabular}{cc|cc|cc|cc|c}
\hline
  \multirow{2}{*}{\textbf{$g_a,g_s$}}~&~~  \multirow{2}{*}{\textbf{$h_a,h_s$}}&\multicolumn{2}{c|}{\textbf{Classification}} & \multicolumn{2}{c|}{\textbf{Detection}} & \multicolumn{2}{c|}{\textbf{Segmentation}}& Params  \\
    &     & BD-rate$\downarrow$ & ~BD-acc$\uparrow$~ & BD-rate$\downarrow$ & BD-mAP$\uparrow$ & BD-rate$\downarrow$ & BD-mAP$\uparrow$ &(M) \\

\hline
 \rowcolor{red!10} 
 \checkmark& & -82.00\% & 18.71 & -56.17\%&3.84&-52.65\%&3.17&0.28\\
& \checkmark & -1.56\% & 0.27 & -1.32\%& 0.07& -1.11\%&0.09 &0.26 \\
\checkmark& \checkmark & -81.27\% & 18.80 & -56.94\%& 3.91& -53.21\%&3.19&0.55 \\

\hline
\end{tabular}
\label{tab:appendix-on-entropy-model}
\vspace{-10pt}
\end{table}
\subsection{Compared with Naive Adapter}

We also replace our SFMA with the naive adapter in ~\cite{chen2022adaptformer} to perform feature adaptation. The naive adapter consists of two linear layers with a ReLU activation layer, which is limited in achieving spatial and frequency modulation like our SFMA.
\cref{tab:appendix:naive}
shows that the naive adapter is inferior to our proposed SFA and FMA under similar trainable parameters, demonstrating the effectiveness of our SFMA.

\begin{table}[htp]
\vspace{-10pt}
\centering
\caption{Ablations on using naive adapter. Naive-$n$ denotes that the naive adapter with the middle dimension set to $n$.}
\begin{tabular}{l|cc|cc|cc|c}
\hline
  \multirow{2}{*}{\textbf{Method}} & \multicolumn{2}{c|}{\textbf{Classification}} & \multicolumn{2}{c|}{\textbf{Detection}} & \multicolumn{2}{c|}{\textbf{Segmentation}}& Params  \\
         & BD-rate$\downarrow$ & ~BD-acc$\uparrow$~ & BD-rate$\downarrow$ & BD-mAP$\uparrow$ & BD-rate$\downarrow$ & BD-mAP$\uparrow$ &(M) \\

\hline
 \rowcolor{red!10} 
SFMA & -82.00\% & 18.71 & -56.17\%&3.84&-52.65\%&3.17&0.28\\
SMA-only & -73.82\% & 16.30 & -51.94\%& 3.61& -48.16\%&2.92 &0.16 \\
FMA-only & -77.40\% & 17.29 & -52.86\%& 3.32& -49.58\%&2.90 &0.12 \\
Naive-64 & -69.72\% & 15.61 & -46.83\%& 2.82& -42.34\% &2.35&0.11 \\
Naive-96 & -71.23\% & 15.92 & -47.92\%& 2.94&  -44.17\% &2.47&0.16 \\

\hline
\end{tabular}
\label{tab:appendix:naive}
\vspace{-10pt}
\end{table}

\section{More Qualitative Results}
\label{sectionJ}
We  provide additional qualitative results for detection and segmentation tasks. These qualitative results further demonstrate the superiority of our framework, which can effectively reduce the latent redundancies for machine vision, thus achieving better task performance with lower bitrates.
\begin{figure*}[h]
\vspace{0pt}
\begin{center}
\subfloat{\includegraphics[width=.99\linewidth]{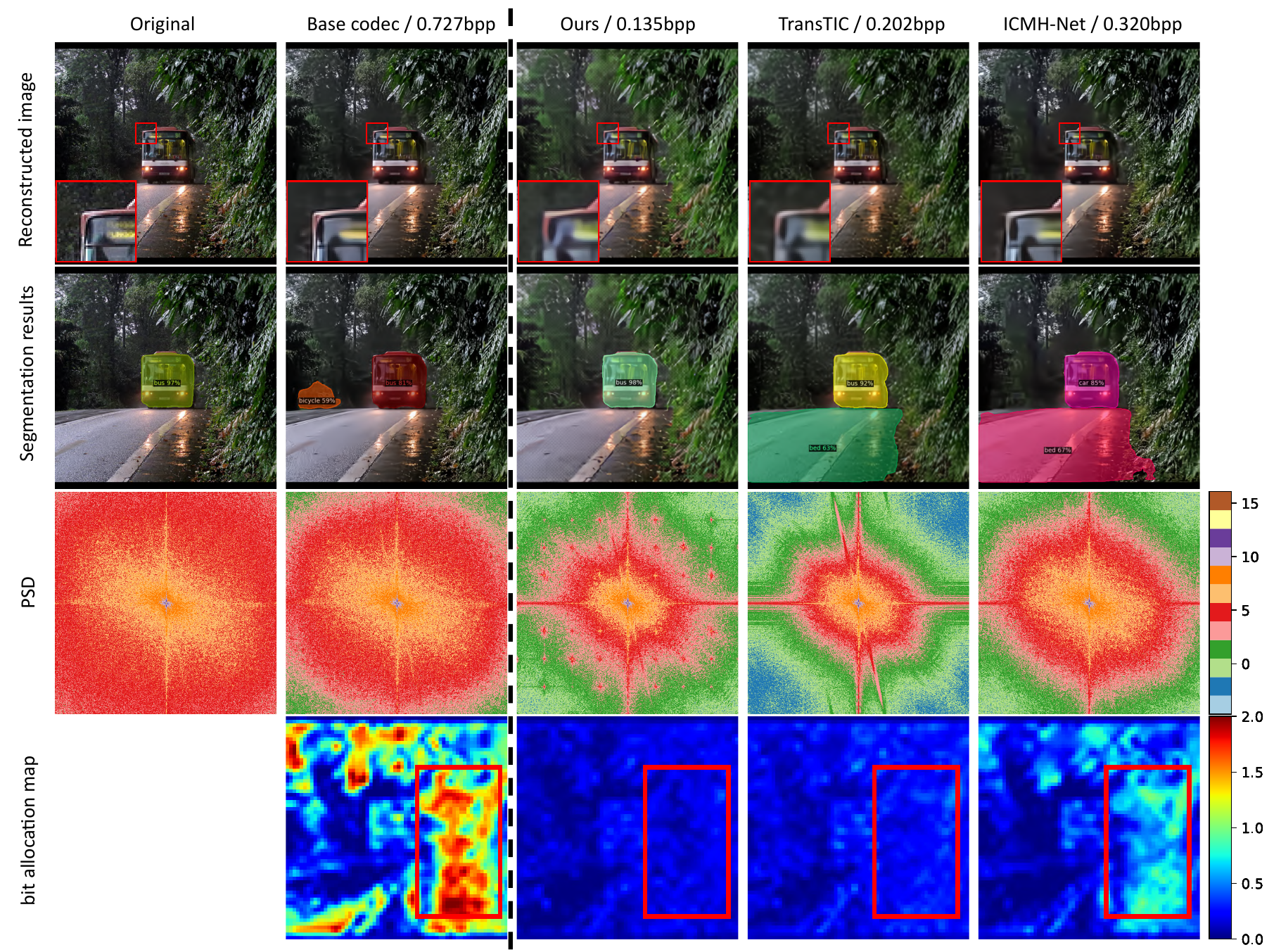}}
\caption{Qualitative comparison of our Adapt-ICMH with other ICMH methods. \textbf{First row:} The original image and decoded image of each method. We show the decoded images for machine vision of three ICMH methods (left). \textbf{Second row:} The object detection results of each image. \textbf{Third row:} The log power spectral density maps of each image. \textbf{Bottom row:} The bit allocation maps for $\bm{\hat{y}}$ of each method.  }
\label{fig:appendix-seg}
\end{center}
\vspace{-10pt}
\end{figure*}
\begin{figure*}[h]
\vspace{0pt}
\begin{center}
\subfloat{\includegraphics[width=.99\linewidth]{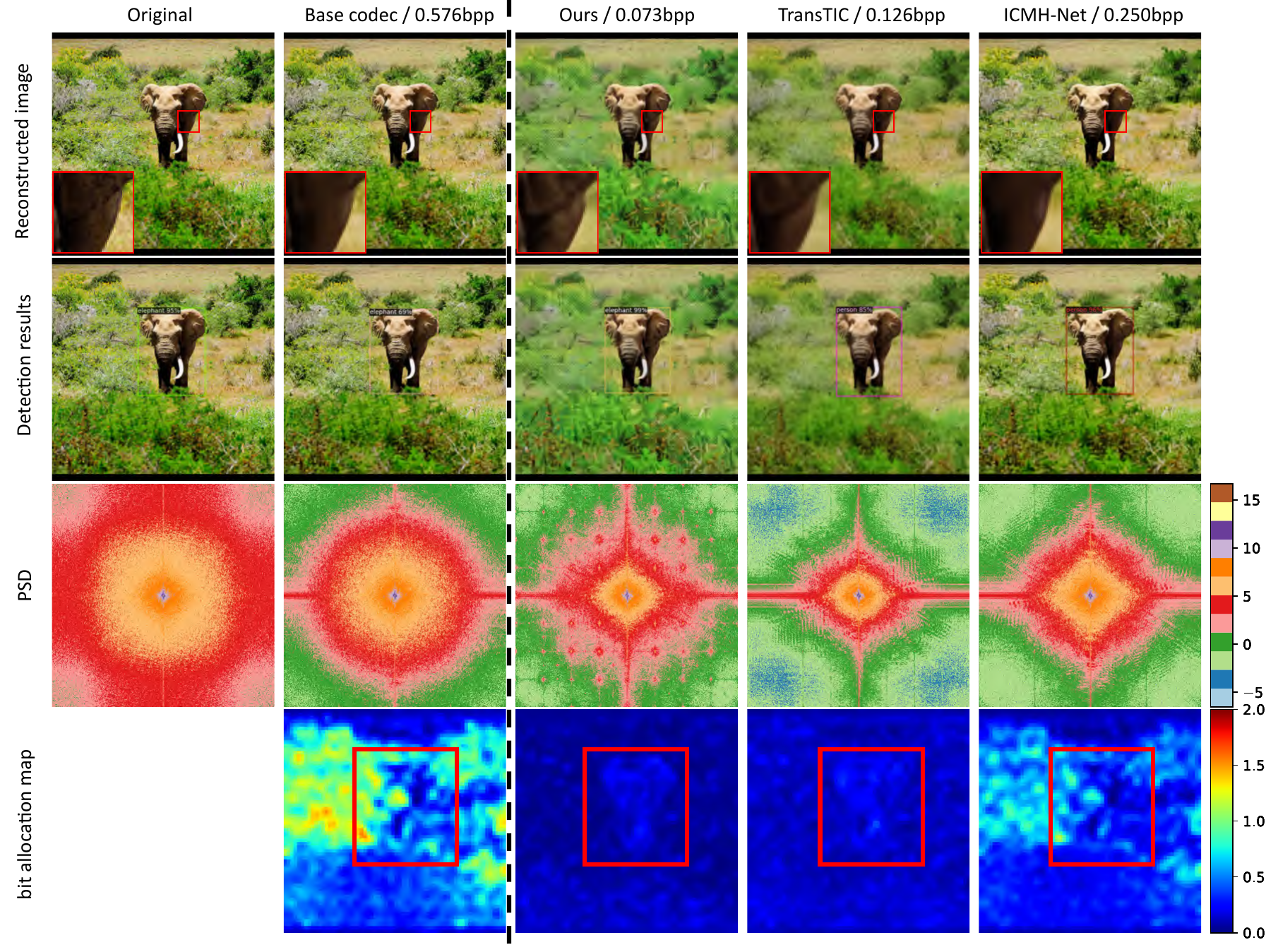}}
\caption{Qualitative comparison of our Adapt-ICMH with other ICMH methods. \textbf{First row:} The original image and decoded image of each method. We show the decoded images for machine vision of three ICMH methods (left). \textbf{Second row:} The instance segmentation results of each image. \textbf{Third row:} The log power spectral density maps of each image. \textbf{Bottom row:} The bit allocation maps for $\bm{\hat{y}}$ of each method.  }
\label{fig:appendix-seg}
\end{center}
\vspace{-10pt}
\end{figure*}

\clearpage

\section{Pytorch Implementation of SFMA}
\label{sectionK}
\begin{lstlisting}[language=Python, caption={Pytorch implementation of SFMA}, label=lst:sfma]
class SFMA(nn.Module):
    def __init__(self, in_dim=128, middle_dim=64,factor=1):
        super().__init__()
        self.s_down1 = nn.Conv2d(in_dim, middle_dim, 1, 1, 0)
        self.s_down2 = nn.Conv2d(in_dim, middle_dim, 1, 1, 0)
        self.s_dw = nn.Conv2d(middle_dim, middle_dim, 5, 1, 2, groups=middle_dim)
        self.s_relu = nn.ReLU(inplace=True)
        self.s_up = nn.Conv2d(middle_dim, in_dim, 1, 1, 0)
       
        self.f_down = nn.Conv2d(in_dim, middle_dim, 1, 1, 0)
        self.f_relu1 = nn.ReLU(inplace=True)
        self.f_relu2 = nn.ReLU(inplace=True)
        self.f_up = nn.Conv2d(middle_dim, in_dim, 1, 1, 0)
        self.f_dw = nn.Conv2d(middle_dim, middle_dim, 3, 1, 1, groups=middle_dim)
        self.f_inter = nn.Conv2d(middle_dim, middle_dim, 1, 1, 0)
        self.sg = nn.Sigmoid()
    
    def forward(self, x):
        '''
        input: 
        x: intermediate feature 
        output:
        x_tilde: adapted feature
        '''
        _, _, H, W = x.shape

        y = torch.fft.rfft2(self.f_down(x), dim=(2, 3), norm='backward')
        y_amp = torch.abs(y)
        y_phs = torch.angle(y)
        y_amp_modulation = self.f_inter(self.f_relu1(self.f_dw(y_amp)))
        y_amp = y_amp * self.sg(y_amp_modulation)
        y_real = y_amp * torch.cos(y_phs)
        y_img = y_amp * torch.sin(y_phs)
        y = torch.complex(y_real, y_img)
        y = torch.fft.irfft2(y, s=(H, W), norm='backward')
        
        f_modulate = self.f_up(self.f_relu2(y))
        s_modulate = self.s_up(self.s_relu(self.s_dw(self.s_down1(x)) * self.s_down2(x)))
        x_tilde = x + (s_modulate + f_modulate)*factor
        return x_tilde 
\end{lstlisting}
\section{Limitation and Future Work}
\label{sectionL}
A potential limitation of our Adapt-ICMH is that it cannot directly achieve scalable coding for machine and human vision. However, our proposed SFMA can incorporate existing scalable coding ICMH frameworks~\cite{liu2023icmh} and boost their performance, as demonstrated in our paper.  Further, we will extend our method into more machine vision tasks, such as pose estimation~\cite{sun2019deep,cao2017realtime,li2021hierarchical,li2023pose,zheng2023actionprompt}, person re-identification~\cite{ye2021deep,somers2023body,yan2023clip}. Additionally, while we only focus on image compression in this paper, video coding for machines (VCM)~\cite{duan2020video,yang2024video} is also a current topic of interest. In our future work,  we aim to expand the scope of SFMA to encompass VCM to further demonstrate the superiority of our propose framework.

\bibliographystyle{splncs04}
\bibliography{egbib}

\begin{thebibliography}{10}
\providecommand{\url}[1]{\texttt{#1}}
\providecommand{\urlprefix}{URL }
\providecommand{\doi}[1]{https://doi.org/#1}

\bibitem{bai2022towards}
Bai, Y., Yang, X., Liu, X., Jiang, J., Wang, Y., Ji, X., Gao, W.: Towards end-to-end image compression and analysis with transformers. In: AAAI. vol.~36, pp. 104--112 (2022)

\bibitem{balle2020nonlinear}
Ball{\'e}, J., Chou, P.A., Minnen, D., Singh, S., Johnston, N., Agustsson, E., Hwang, S.J., Toderici, G.: Nonlinear transform coding. IEEE JSTSP  \textbf{15}(2),  339--353 (2020)

\bibitem{balle2015density}
Ball{\'e}, J., Laparra, V., Simoncelli, E.P.: Density modeling of images using a generalized normalization transformation. In: ICLR (2016)

\bibitem{balle2016end}
Ball{\'e}, J., Laparra, V., Simoncelli, E.P.: End-to-end optimized image compression. In: ICLR (2016)

\bibitem{balle2018variational}
Ball{\'e}, J., Minnen, D., Singh, S., Hwang, S.J., Johnston, N.: Variational image compression with a scale hyperprior. In: ICLR (2018)

\bibitem{begaint2020compressai}
B{\'e}gaint, J., Racap{\'e}, F., Feltman, S., Pushparaja, A.: Compressai: a pytorch library and evaluation platform for end-to-end compression research. arXiv preprint arXiv:2011.03029  (2020)

\bibitem{bdrate}
Bjontegaard, G.: Calculation of average psnr differences between rd-curves. In: VCEG-M33 (2001)

\bibitem{campos2019content}
Campos, J., Meierhans, S., Djelouah, A., Schroers, C.: Content adaptive optimization for neural image compression. In: CVPRW (2019)

\bibitem{cao2017realtime}
Cao, Z., Simon, T., Wei, S.E., Sheikh, Y.: Realtime multi-person 2d pose estimation using part affinity fields. In: Proceedings of the IEEE conference on computer vision and pattern recognition. pp. 7291--7299 (2017)

\bibitem{carion2020end}
Carion, N., Massa, F., Synnaeve, G., Usunier, N., Kirillov, A., Zagoruyko, S.: End-to-end object detection with transformers. In: ECCV. pp. 213--229. Springer (2020)

\bibitem{chen2022adaptformer}
Chen, S., Ge, C., Tong, Z., Wang, J., Song, Y., Wang, J., Luo, P.: Adaptformer: Adapting vision transformers for scalable visual recognition. In: NeurIPS. vol.~35, pp. 16664--16678 (2022)

\bibitem{chen2023transtic}
Chen, Y.H., Weng, Y.C., Kao, C.H., Chien, C., Chiu, W.C., Peng, W.H.: Transtic: Transferring transformer-based image compression from human perception to machine perception. In: ICCV. pp. 23297--23307 (2023)

\bibitem{chen2022vision}
Chen, Z., Duan, Y., Wang, W., He, J., Lu, T., Dai, J., Qiao, Y.: Vision transformer adapter for dense predictions. In: ICLR (2023)

\bibitem{cheng2020learned}
Cheng, Z., Sun, H., Takeuchi, M., Katto, J.: Learned image compression with discretized gaussian mixture likelihoods and attention modules. In: CVPR. pp. 7939--7948 (2020)

\bibitem{choi2022scalable}
Choi, H., Baji{\'c}, I.V.: Scalable image coding for humans and machines. IEEE TIP  \textbf{31},  2739--2754 (2022)

\bibitem{chollet2017xception}
Chollet, F.: Xception: Deep learning with depthwise separable convolutions. In: CVPR. pp. 1251--1258 (2017)

\bibitem{codevilla2021learned}
Codevilla, F., Simard, J.G., Goroshin, R., Pal, C.: Learned image compression for machine perception. arXiv preprint arXiv:2111.02249  (2021)

\bibitem{deng2009imagenet}
Deng, J., Dong, W., Socher, R., Li, L.J., Li, K., Fei-Fei, L.: Imagenet: A large-scale hierarchical image database. In: CVPR. pp. 248--255 (2009)

\bibitem{ding2022motion}
Ding, S., Li, M., Yang, T., Qian, R., Xu, H., Chen, Q., Wang, J., Xiong, H.: Motion-aware contrastive video representation learning via foreground-background merging. In: Proceedings of the IEEE/CVF conference on computer vision and pattern recognition. pp. 9716--9726 (2022)

\bibitem{dosovitskiy2020image}
Dosovitskiy, A., Beyer, L., Kolesnikov, A., Weissenborn, D., Zhai, X., Unterthiner, T., Dehghani, M., Minderer, M., Heigold, G., Gelly, S., et~al.: An image is worth 16x16 words: Transformers for image recognition at scale. In: ICLR (2020)

\bibitem{duan2020video}
Duan, L., Liu, J., Yang, W., Huang, T., Gao, W.: Video coding for machines: A paradigm of collaborative compression and intelligent analytics. IEEE Transactions on Image Processing  \textbf{29},  8680--8695 (2020)

\bibitem{feng2023semantically}
Feng, R., Gao, Y., Jin, X., Feng, R., Chen, Z.: Semantically structured image compression via irregular group-based decoupling. In: ICCV (2023)

\bibitem{feng2022image}
Feng, R., Jin, X., Guo, Z., Feng, R., Gao, Y., He, T., Zhang, Z., Sun, S., Chen, Z.: Image coding for machines with omnipotent feature learning. In: ECCV. pp. 510--528. Springer (2022)

\bibitem{feng2023prompt}
Feng, R., Liu, J., Jin, X., Pan, X., Sun, H., Chen, Z.: Prompt-icm: A unified framework towards image coding for machines with task-driven prompts. arXiv preprint arXiv:2305.02578  (2023)

\bibitem{fischer2022boosting}
Fischer, K., Brand, F., Kaup, A.: Boosting neural image compression for machines using latent space masking. IEEE TCSVT  (2022)

\bibitem{girshick2015fast}
Girshick, R.: Fast r-cnn. In: Proceedings of the IEEE international conference on computer vision. pp. 1440--1448 (2015)

\bibitem{he2021towards}
He, J., Zhou, C., Ma, X., Berg-Kirkpatrick, T., Neubig, G.: Towards a unified view of parameter-efficient transfer learning. In: ICLR (2022)

\bibitem{he2017mask}
He, K., Gkioxari, G., Doll{\'a}r, P., Girshick, R.: Mask r-cnn. In: Proceedings of the IEEE international conference on computer vision. pp. 2961--2969 (2017)

\bibitem{he2016deep}
He, K., Zhang, X., Ren, S., Sun, J.: Deep residual learning for image recognition. In: CVPR. pp. 770--778 (2016)

\bibitem{he2023parameter}
He, X., Li, C., Zhang, P., Yang, J., Wang, X.E.: Parameter-efficient model adaptation for vision transformers. In: AAAI. vol.~37, pp. 817--825 (2023)

\bibitem{houlsby2019parameter}
Houlsby, N., Giurgiu, A., Jastrzebski, S., Morrone, B., De~Laroussilhe, Q., Gesmundo, A., Attariyan, M., Gelly, S.: Parameter-efficient transfer learning for nlp. In: ICML. pp. 2790--2799. PMLR (2019)

\bibitem{huang2017densely}
Huang, G., Liu, Z., Van Der~Maaten, L., Weinberger, K.Q.: Densely connected convolutional networks. In: Proceedings of the IEEE conference on computer vision and pattern recognition. pp. 4700--4708 (2017)

\bibitem{jia2022visual}
Jia, M., Tang, L., Chen, B.C., Cardie, C., Belongie, S., Hariharan, B., Lim, S.N.: Visual prompt tuning. In: ECCV. pp. 709--727. Springer (2022)

\bibitem{khattak2023maple}
Khattak, M.U., Rasheed, H., Maaz, M., Khan, S., Khan, F.S.: Maple: Multi-modal prompt learning. In: CVPR. pp. 19113--19122 (2023)

\bibitem{koyuncu2022contextformer}
Koyuncu, A.B., Gao, H., Boev, A., Gaikov, G., Alshina, E., Steinbach, E.: Contextformer: A transformer with spatio-channel attention for context modeling in learned image compression. In: ECCV (2022)

\bibitem{lee2018context}
Lee, J., Cho, S., Beack, S.K.: Context-adaptive entropy model for end-to-end optimized image compression. In: ICLR (2019)

\bibitem{lester2021power}
Lester, B., Al-Rfou, R., Constant, N.: The power of scale for parameter-efficient prompt tuning. In: EMNLP (2021)

\bibitem{li2023frequency}
Li, H., Li, S., Dai, W., Li, C., Zou, J., Xiong, H.: Frequency-aware transformer for learned image compression. In: The Twelfth International Conference on Learning Representations (2024), \url{https://openreview.net/forum?id=HKGQDDTuvZ}

\bibitem{li2021hierarchical}
Li, H., Shi, B., Dai, W., Chen, Y., Wang, B., Sun, Y., Guo, M., Li, C., Zou, J., Xiong, H.: Hierarchical graph networks for 3d human pose estimation. arXiv preprint arXiv:2111.11927  (2021)

\bibitem{li2023pose}
Li, H., Shi, B., Dai, W., Zheng, H., Wang, B., Sun, Y., Guo, M., Li, C., Zou, J., Xiong, H.: Pose-oriented transformer with uncertainty-guided refinement for 2d-to-3d human pose estimation. In: AAAI. vol.~37, pp. 1296--1304 (2023)

\bibitem{lin2014microsoft}
Lin, T.Y., Maire, M., Belongie, S., Hays, J., Perona, P., Ramanan, D., Doll{\'a}r, P., Zitnick, C.L.: Microsoft coco: Common objects in context. In: ECCV. pp. 740--755 (2014)

\bibitem{liu2024rate}
Liu, J., Feng, R., Qi, Y., Chen, Q., Chen, Z., Zeng, W., Jin, X.: Rate-distortion-cognition controllable versatile neural image compression. In: ECCV. Springer (2024)

\bibitem{liu2023composable}
Liu, J., Jin, X., Feng, R., Chen, Z., Zeng, W.: Composable image coding for machine via task-oriented internal adaptor and external prior. In: VCIP. pp.~1--5 (2023)

\bibitem{liu2022improving}
Liu, J., Sun, H., Katto, J.: Improving multiple machine vision tasks in the compressed domain. In: ICPR. pp. 331--337. IEEE (2022)

\bibitem{liu2023learned}
Liu, J., Sun, H., Katto, J.: Learned image compression with mixed transformer-cnn architectures. In: CVPR. pp. 14388--14397 (2023)

\bibitem{liu2021semantics}
Liu, K., Liu, D., Li, L., Yan, N., Li, H.: Semantics-to-signal scalable image compression with learned revertible representations. IJCV  \textbf{129}(9),  2605--2621 (2021)

\bibitem{liu2023icmh}
Liu, L., Hu, Z., Chen, Z., Xu, D.: Icmh-net: Neural image compression towards both machine vision and human vision. In: ACM MM. pp. 8047--8056 (2023)

\bibitem{liu2023pre}
Liu, P., Yuan, W., Fu, J., Jiang, Z., Hayashi, H., Neubig, G.: Pre-train, prompt, and predict: A systematic survey of prompting methods in natural language processing. ACM Computing Surveys  \textbf{55}(9),  1--35 (2023)

\bibitem{liu2021swin}
Liu, Z., Lin, Y., Cao, Y., Hu, H., Wei, Y., Zhang, Z., Lin, S., Guo, B.: Swin transformer: Hierarchical vision transformer using shifted windows. In: ICCV. pp. 10012--10022 (2021)

\bibitem{lu2022transformer}
Lu, M., Guo, P., Shi, H., Cao, C., Ma, Z.: Transformer-based image compression. In: DCC. pp. 469--469 (2022)

\bibitem{lv2023dynamic}
Lv, Y., Xiang, J., Zhang, J., Yang, W., Han, X., Yang, W.: Dynamic low-rank instance adaptation for universal neural image compression. In: ACM MM. pp. 632--642 (2023)

\bibitem{mentzer2019practical}
Mentzer, F., Agustsson, E., Tschannen, M., Timofte, R., Gool, L.V.: Practical full resolution learned lossless image compression. In: CVPR. pp. 10629--10638 (2019)

\bibitem{minnen2018joint}
Minnen, D., Ball{\'e}, J., Toderici, G.D.: Joint autoregressive and hierarchical priors for learned image compression. In: NeurIPS. vol.~31 (2018)

\bibitem{minnen2020channel}
Minnen, D., Singh, S.: Channel-wise autoregressive entropy models for learned image compression. In: ICIP. pp. 3339--3343 (2020)

\bibitem{nair2010rectified}
Nair, V., Hinton, G.E.: Rectified linear units improve restricted boltzmann machines. In: ICML. pp. 807--814 (2010)

\bibitem{pfeiffer2020adapterfusion}
Pfeiffer, J., Kamath, A., R{\"u}ckl{\'e}, A., Cho, K., Gurevych, I.: Adapterfusion: Non-destructive task composition for transfer learning. arXiv preprint arXiv:2005.00247  (2020)

\bibitem{pfeiffer2020adapterhub}
Pfeiffer, J., R{\"u}ckl{\'e}, A., Poth, C., Kamath, A., Vuli{\'c}, I., Ruder, S., Cho, K., Gurevych, I.: Adapterhub: A framework for adapting transformers. In: EMNLP. pp. 46--54 (2020)

\bibitem{qian2021entroformer}
Qian, Y., Sun, X., Lin, M., Tan, Z., Jin, R.: Entroformer: A transformer-based entropy model for learned image compression. In: ICLR (2022)

\bibitem{ren2015faster}
Ren, S., He, K., Girshick, R., Sun, J.: Faster r-cnn: Towards real-time object detection with region proposal networks. In: NeurIPS. vol.~28 (2015)

\bibitem{shen2023dec}
Shen, S., Yue, H., Yang, J.: Dec-adapter: Exploring efficient decoder-side adapter for bridging screen content and natural image compression. In: CVPR. pp. 12887--12896 (2023)

\bibitem{shi2022transformer}
Shi, B., Jiang, D., Zhang, X., Li, H., Dai, W., Zou, J., Xiong, H., Tian, Q.: A transformer-based decoder for semantic segmentation with multi-level context mining. In: European Conference on Computer Vision. pp. 624--639. Springer (2022)

\bibitem{somers2023body}
Somers, V., De~Vleeschouwer, C., Alahi, A.: Body part-based representation learning for occluded person re-identification. In: Proceedings of the IEEE/CVF winter conference on applications of computer vision. pp. 1613--1623 (2023)

\bibitem{strudel2021segmenter}
Strudel, R., Garcia, R., Laptev, I., Schmid, C.: Segmenter: Transformer for semantic segmentation. In: Proceedings of the IEEE/CVF international conference on computer vision. pp. 7262--7272 (2021)

\bibitem{sun2019deep}
Sun, K., Xiao, B., Liu, D., Wang, J.: Deep high-resolution representation learning for human pose estimation. In: CVPR. pp. 5693--5703 (2019)

\bibitem{tsubota2023universal}
Tsubota, K., Akutsu, H., Aizawa, K.: Universal deep image compression via content-adaptive optimization with adapters. In: WACV. pp. 2529--2538 (2023)

\bibitem{wang2023adapting}
Wang, Y., Shi, B., Zhang, X., Li, J., Liu, Y., Dai, W., Li, C., Xiong, H., Tian, Q.: Adapting shortcut with normalizing flow: An efficient tuning framework for visual recognition. In: CVPR. pp. 15965--15974 (2023)

\bibitem{xie2021segformer}
Xie, E., Wang, W., Yu, Z., Anandkumar, A., Alvarez, J.M., Luo, P.: Segformer: Simple and efficient design for semantic segmentation with transformers. In: NeurIPS. vol.~34, pp. 12077--12090 (2021)

\bibitem{yan2023clip}
Yan, S., Dong, N., Zhang, L., Tang, J.: Clip-driven fine-grained text-image person re-identification. IEEE Transactions on Image Processing  (2023)

\bibitem{yang2021towards}
Yang, S., Hu, Y., Yang, W., Duan, L.Y., Liu, J.: Towards coding for human and machine vision: Scalable face image coding. IEEE TMM  \textbf{23},  2957--2971 (2021)

\bibitem{yang2024video}
Yang, W., Huang, H., Hu, Y., Duan, L.Y., Liu, J.: Video coding for machines: Compact visual representation compression for intelligent collaborative analytics. IEEE Transactions on Pattern Analysis and Machine Intelligence  (2024)

\bibitem{ye2021deep}
Ye, M., Shen, J., Lin, G., Xiang, T., Shao, L., Hoi, S.C.: Deep learning for person re-identification: A survey and outlook. IEEE transactions on pattern analysis and machine intelligence  \textbf{44}(6),  2872--2893 (2021)

\bibitem{zheng2023actionprompt}
Zheng, H., Li, H., Shi, B., Dai, W., Wang, B., Sun, Y., Guo, M., Xiong, H.: Actionprompt: Action-guided 3d human pose estimation with text and pose prompting. In: 2023 IEEE International Conference on Multimedia and Expo (ICME). pp. 2657--2662. IEEE (2023)

\bibitem{zheng2024bem}
Zheng, H., Zhou, L., Li, H., Su, J., Wei, X., Xu, X.: Bem: Balanced and entropy-based mix for long-tailed semi-supervised learning. In: Proceedings of the IEEE/CVF Conference on Computer Vision and Pattern Recognition (CVPR). pp. 22893--22903 (June 2024)

\bibitem{zheng2021rethinking}
Zheng, S., Lu, J., Zhao, H., Zhu, X., Luo, Z., Wang, Y., Fu, Y., Feng, J., Xiang, T., Torr, P.H., et~al.: Rethinking semantic segmentation from a sequence-to-sequence perspective with transformers. In: Proceedings of the IEEE/CVF conference on computer vision and pattern recognition. pp. 6881--6890 (2021)

\bibitem{zhu2020deformable}
Zhu, X., Su, W., Lu, L., Li, B., Wang, X., Dai, J.: Deformable detr: Deformable transformers for end-to-end object detection. arXiv preprint arXiv:2010.04159  (2020)

\bibitem{zou2022devil}
Zou, R., Song, C., Zhang, Z.: The devil is in the details: Window-based attention for image compression. In: CVPR. pp. 17492--17501 (2022)

\end{thebibliography}

\end{document}